\pdfoutput=1
\documentclass[11pt]{article}
\usepackage[]{acl}
\usepackage{times}
\usepackage{latexsym}
\usepackage[T1]{fontenc}
\usepackage[utf8]{inputenc}
\usepackage{microtype}
\usepackage{inconsolata}
\usepackage{booktabs}
\usepackage{graphicx}
\usepackage{makecell}
\usepackage{multirow}
\usepackage{colortbl}
\usepackage{float}
\usepackage{xcolor}
\usepackage{amsmath}
\usepackage{amssymb}
\usepackage{textcomp}
\usepackage{pifont}
\usepackage{cancel}
\usepackage{enumitem}
\usepackage{courier}
\usepackage{fontawesome5}

\definecolor{blue}{RGB}{50,80,130}
\definecolor{red}{RGB}{120,30,35}
\definecolor{green}{RGB}{34,85,50}
\definecolor{gray}{RGB}{90,100,110}
\definecolor{cyan}{RGB}{70,120,140}
\definecolor{purple}{RGB}{150,110,140}
\definecolor{softred}{RGB}{200,90,95}
\definecolor{softgreen}{RGB}{85,150,100}

\title{
A Large-Scale Multi-Dimensional Empirical Study of LLMs for \\ Conversation Summarization
}

\author{Weixiao Zhou\textsuperscript{\textnormal{$\alpha$}} \\
		\vspace{5pt}
		\textbf{Gengyao Li}\textsuperscript{\textnormal{$\gamma$}} \enspace
		\textbf{Junnan Zhu}\textsuperscript{\textnormal{$\beta$}}\thanks{Corresponding Author} \enspace
		\textbf{Xianfu Cheng}\textsuperscript{\textnormal{$\alpha$}} \enspace
		\textbf{Feifei Zhai}\textsuperscript{\textnormal{$\beta\gamma$}} \enspace
		\textbf{Zhoujun Li}\textsuperscript{\textnormal{$\alpha$}}\footnotemark[\value{footnote}] \\
		\textsuperscript{\textnormal{$\alpha$}}CCSE, Beihang University \quad
		\textsuperscript{\textnormal{$\beta$}}MAIS, CASIA \quad
		\textsuperscript{\textnormal{$\gamma$}}Fanyu AI Laboratory \smallskip \\
        {\fontsize{11.5pt}{0pt}\selectfont \texttt{wxzhou@buaa.edu.cn}}}
\begin{document}
\maketitle

\begin{abstract}
Despite the significant advancement of LLMs in conversation summarization, their evaluation remains limited by insufficient scenarios, input lengths, and sample sizes. Furthermore, existing benchmarks often omit frontier reasoning systems and efficient small models, or lack fine-grained, multi-dimensional assessments. To bridge these gaps, we propose \textsc{OmniCSEval}, a unified benchmark comprising 1,800 diverse conversations across six real-world scenarios, featuring context lengths ranging from 128 to 32k tokens. For fine-grained evaluation, we employ a bidirectional fact-checking framework that integrates key fact matching to assess completeness and conciseness, alongside summary fact verification to evaluate faithfulness. To ensure reliable assessment, we establish a human-LLM collaborative pipeline for key fact extraction and a multi-LLM consensus verifier for summary fact decomposition. Leveraging this framework, we evaluate 28 LLMs across four distinct categories grouped by reasoning capability and model scale. Our extensive empirical study reveals critical insights regarding the cross-scenario challenges current LLMs continue to face, the impacts of reasoning and scale, and the efficiency and adaptability of reasoning models. We also provide guidance for system selection in real-world deployments\footnote{\fontsize{8.1pt}{0pt}\selectfont \url{https://github.com/zhouweixiao/OmniCSEval}.}.
\end{abstract}

\section{Introduction}
\vspace{-2pt}
Conversation summarization aims to distill key information from a dialogue into a faithful summary \citep{10.1613/jair.1.16674}. The advancement of LLMs has significantly improved this field \citep{zhou-etal-2023-multi, zhu-etal-2025-factual}. However, existing evaluation studies remain constrained by narrow conversation scenarios \citep{zhu-etal-2023-annotating, gao-wan-2022-dialsummeval}, insufficient sample sizes \citep{lee-etal-2024-unisumeval, min-etal-2025-towards}, and context length restrictions \citep{laban-etal-2023-summedits, tang-etal-2024-tofueval}. In Table~\ref{tab:benchmark_comparison}, we present a comprehensive comparison of current conversation summarization benchmarks.

\begin{figure}[t]
  \centering
  \setlength{\abovecaptionskip}{10pt}
  \includegraphics[width=0.99999\linewidth]{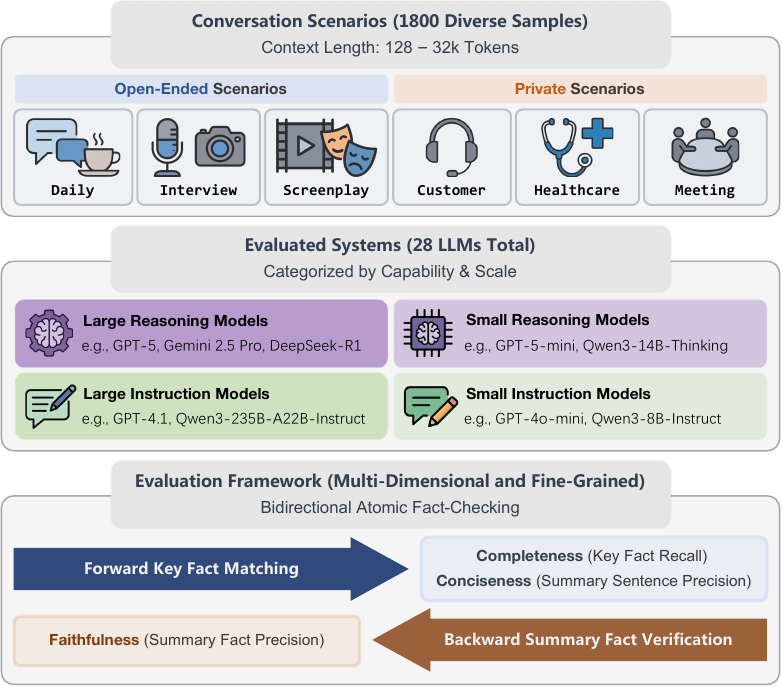}
  \caption{Core components of \textsc{OmniCSEval}. We scale the diversity of conversations and evaluate LLMs with a fine-grained and interpretable evaluation framework.}
  \label{fig:introduction}
  \vspace{0pt}
\end{figure}

Moreover, reasoning language models (RLMs) demonstrate exceptional capabilities in complex logical tasks by scaling inference compute \citep{li202512surveyreasoning, chen2025reasoningerasurveylong}, while small language models\footnote{In this study, \textit{small} models are defined as open-weight LLMs with fewer than 15B parameters (e.g., Qwen3-8B, -14B) or lightweight variants of proprietary LLMs (e.g., GPT-5-nano, -mini). See Table~\ref{tab:evaluated_llm_details} (Appendix~\ref{sec:evaluated_llm_details}) for our detailed taxonomy.} (SLMs) \citep{subramanian2025smalllanguagemodelsslms, 10.1145/3768165} are increasingly vital for cost-effective applications. Nevertheless, the impacts of explicit reasoning and model scale on conversation summarization remain underexplored. While recent works \citep{jin2025reasoningnotcomprehensiveevaluation, yuan2025understandingllmreasoningabstractive, fu-etal-2024-tiny, xu-etal-2025-evaluating} offer preliminary investigations into these types of models, they lack fine-grained evaluation on multiple dimensions \citep{song-etal-2024-finesure}, relying heavily on coarse similarity-based metrics \citep{lin2004rouge, zhang2019bertscore}.

\defcitealias{gao-wan-2022-dialsummeval}{DialSummEval}
\defcitealias{zhu-etal-2023-annotating}{DiaSumFact}
\defcitealias{laban-etal-2023-summedits}{SummEdits}
\defcitealias{tang-etal-2024-tofueval}{TofuEval}
\defcitealias{lee-etal-2024-unisumeval}{UniSumEval}
\defcitealias{min-etal-2025-towards}{MSumBench}

\definecolor{MilkRed}{HTML}{F2B6B6}
\definecolor{MilkGreen}{HTML}{C5E0B4}

\begin{table*}[t]
    \centering
    \small
    \setlength{\abovecaptionskip}{8pt}
    \setlength{\tabcolsep}{3.5pt}
    \resizebox{\linewidth}{!}{
    \begin{tabular}{lccccccccc}
        \toprule[1pt]
        \textbf{Benchmark} & \textbf{\makecell{Dialogue \\ Scenarios}} & \textbf{\makecell{Context \\ Range}} & \textbf{\makecell{Dialogue \\ Samples}} & \textbf{\makecell{Evaluated \\ LLMs}} & \textbf{\makecell{Reasoning \\ LLMs}} & \textbf{\makecell{Small \\ LLMs}} & \textbf{\makecell{Evaluation \\ Dimensions}} & \textbf{\makecell{Evaluation \\ Granularity}} & \textbf{\makecell{Scoring \\ Method}} \\
        \midrule[0.5pt]
        \citetalias{gao-wan-2022-dialsummeval} & \cellcolor{MilkRed!70}1 (1 Source) & \cellcolor{MilkRed!70}24 -- 492 & \cellcolor{MilkRed!70}100 & \cellcolor{MilkRed!70}No & \cellcolor{MilkRed!70}No & \cellcolor{MilkRed!70}No & \cellcolor{MilkGreen!70}Multiple & \cellcolor{MilkRed!70}Summary & \cellcolor{MilkRed!70}Likert \\
        \citetalias{zhu-etal-2023-annotating} & \cellcolor{MilkRed!70}2 (2 Sources) & \cellcolor{MilkRed!70}23 -- 599 & \cellcolor{MilkRed!70}120 & \cellcolor{MilkRed!70}No & \cellcolor{MilkRed!70}No & \cellcolor{MilkRed!70}No & \cellcolor{MilkRed!70}Single & \cellcolor{MilkRed!70}Sentence & \cellcolor{MilkGreen!70}Percentage \\
        \citetalias{tang-etal-2024-tofueval} & \cellcolor{MilkRed!70}2 (2 Sources) & \cellcolor{MilkRed!70}746 -- 1.3k & \cellcolor{MilkRed!70}100 & \cellcolor{MilkGreen!70}5 Models & \cellcolor{MilkRed!70}No & \cellcolor{MilkGreen!70}3 Models & \cellcolor{MilkGreen!70}Multiple & \cellcolor{MilkRed!70}Sentence & \cellcolor{MilkGreen!70}Percentage \\
        \citetalias{laban-etal-2023-summedits} & \cellcolor{MilkRed!70}5 (7 Sources) & \cellcolor{MilkRed!70} 41 -- 1.8k & \cellcolor{MilkRed!70}280 & \cellcolor{MilkGreen!70}10 Models & \cellcolor{MilkRed!70}No & \cellcolor{MilkGreen!70}1 Model & \cellcolor{MilkRed!70}Single & \cellcolor{MilkRed!70}Summary & \cellcolor{MilkRed!70}Ternary \\
        \citetalias{lee-etal-2024-unisumeval} & \cellcolor{MilkRed!70}4 (4 Sources) & \cellcolor{MilkRed!70}65 -- 3.5k & \cellcolor{MilkRed!70}100 & \cellcolor{MilkGreen!70}7 Models & \cellcolor{MilkRed!70}No & \cellcolor{MilkGreen!70}3 Models & \cellcolor{MilkGreen!70}Multiple & \cellcolor{MilkGreen!70}Fact \& Sentence & \cellcolor{MilkGreen!70}Percentage \\
        \citetalias{min-etal-2025-towards} & \cellcolor{MilkRed!70}3 (3 Sources) & \cellcolor{MilkRed!70}187 -- 5.9k & \cellcolor{MilkRed!70}75 & \cellcolor{MilkGreen!70}6 Models & \cellcolor{MilkRed!70}No & \cellcolor{MilkRed!70}No & \cellcolor{MilkGreen!70}Multiple & \cellcolor{MilkGreen!70}Fact \& Sentence & \cellcolor{MilkGreen!70}Percentage \\
        \midrule[0.5pt]
       	\textsc{OmniCSEval} & \cellcolor{MilkGreen!70}\textbf{6 (13 Sources)} & \cellcolor{MilkGreen!70}\textbf{128 -- 32k} & \cellcolor{MilkGreen!70}\textbf{1,800} & \cellcolor{MilkGreen!70}\textbf{28 Models} & \cellcolor{MilkGreen!70}\textbf{14 Models} & \cellcolor{MilkGreen!70}\textbf{15 Models} & \cellcolor{MilkGreen!70}\textbf{Multiple} & \cellcolor{MilkGreen!70}\textbf{Bidirectional Fact} & \cellcolor{MilkGreen!70}\textbf{Percentage} \\
        \bottomrule[1pt]
    \end{tabular}
    }
    \caption{Comparison of our \textsc{OmniCSEval} with existing conversation summarization benchmarks.}
    \label{tab:benchmark_comparison}
    \vspace{-3pt}
\end{table*}

To address these limitations, we introduce \textbf{\textsc{OmniCSEval}} (\underline{Omni} \underline{C}onversation \underline{S}ummarization \underline{Eval}uation), aiming to: (1) establish a \textit{comprehensive} and \textit{trustworthy} benchmark for diverse LLM categories in conversation summarization; (2) investigate the specific \textit{challenges} LLMs face across scenarios and the impacts of reasoning and model scale; and (3) analyze the efficiency of reasoning models in translating thinking length into performance and their adaptability to scenario complexity. Figure~\ref{fig:introduction} presents an overview of our benchmark.

To ensure data breadth, we source 1,800 conversations from 13 high-quality dialogue summarization corpora across six major real-world scenarios: \textit{Daily Life}, \textit{Media Interview}, \textit{Screenplay}, \textit{Customer Service}, \textit{Healthcare}, and \textit{Meeting}. These scenarios span diverse structural complexities, ranging from brief dyadic exchanges in Daily Life to extensive multi-party discussions in Meeting, yielding context lengths that extend from 128 to 32k tokens. Using this dataset, we evaluate 50,400 summaries generated by 28 widely-adopted LLMs, categorized by reasoning capability and scale into four representative groups: \textit{Large Reasoning}, \textit{Small Reasoning}, \textit{Large Instruction}\footnote{Conventionally, \textit{instruction} models are considered equivalent to \textit{non-reasoning} models, both referring to models that do not produce explicit intermediate reasoning processes.}, and \textit{Small Instruction} models.

Our evaluation focuses on three core dimensions: \textit{Completeness}, \textit{Conciseness}, and \textit{Faithfulness}. To achieve a fine-grained and interpretable assessment, we employ an automated bidirectional atomic fact-checking framework\footnote{An atomic fact is a short, self-contained information unit conveying a single factual claim \citep{min-etal-2023-factscore}.}. This integrates the complementary processes of key fact matching \citep{song-etal-2024-finesure} and summary fact verification \citep{wan-etal-2024-acueval}. Completeness is measured as the recall of aligned key facts, conciseness is quantified by the density of fact-matched summary sentences, and faithfulness as the precision of supported summary facts. To guarantee reliable evaluation, we implement a rigorous three-stage human-LLM collaborative pipeline to extract gold-standard key facts and a robust multi-LLM consensus filtering mechanism for summary fact decomposition. Furthermore, we isolate the LLMs utilized in the evaluation pipeline from the evaluated ones to minimize potential bias \citep{zhao-etal-2024-comparative}. We also demonstrate our evaluation's stability, judge-agnostic robustness, and strong alignment with human judgments (see \S\ref{sec:meta_evaluation}).

Through extensive empirical analysis, we reveal several critical findings:
\begin{itemize}[leftmargin=20pt, itemsep=0.4pt, topsep=2.6pt]
  \item LLMs face distinct challenges across scenarios. For instance, they exhibit profound vulnerabilities in completeness and conciseness within Meeting and Screenplay, while struggling with faithfulness in Daily Life.
  \item Reasoning is essential for improving completeness and conciseness, but it also introduces an increased risk of hallucinations.
  \item While explicit reasoning can effectively offset parameter deficiencies, model scale remains the decisive factor for multi-dimensional performance and cross-scenario stability.
  \item Models with longer reasoning processes do not necessarily achieve better performance, particularly in completeness and faithfulness.
  \item A notable number of reasoning models fail to dynamically scale their computational budget to match scenario complexity, exhibiting rigid or even inverse trends in thinking lengths.
\end{itemize}

\section{Related Work}
\vspace{-1pt}
\paragraph{Evaluating LLMs for Dialogue Summarization.}
Despite recent efforts to evaluate LLM-based conversation summarization, existing studies are still bottlenecked by constrained scenarios \citep{tang-etal-2023-context, tang-etal-2024-tofueval}, conversation lengths \citep{laban-etal-2023-summedits, zhou-etal-2025-talking}, evaluation scales \citep{lee-etal-2024-unisumeval, min-etal-2025-towards} and dimensions \citep{ramprasad-etal-2024-analyzing, bao-etal-2025-faithbench}, potentially limiting the generalizability and robustness of their results. For instance, TofuEval \citep{tang-etal-2024-tofueval} is restricted to two scenarios, UniSumEval \citep{lee-etal-2024-unisumeval} and MSumBench \citep{min-etal-2025-towards} assess merely 25 instances per domain, and FaithBench \citep{bao-etal-2025-faithbench} solely focuses on the faithfulness dimension. Furthermore, while emerging literature explores the behavior of reasoning \citep{jin2025reasoningnotcomprehensiveevaluation, yuan2025understandingllmreasoningabstractive} and small language models \citep{fu-etal-2024-tiny}, these works still lack trustworthy evaluation protocols \citep{song-etal-2024-finesure}. Therefore, constructing a comprehensive and reliable benchmark is a critical priority.

\vspace{-1pt}
\paragraph{Summarization Evaluation Methods.}
Conventional similarity-based metrics \citep{NEURIPS2021_e4d2b6e6, zhang2019bertscore} correlate poorly with human judgments. NLI-based \citep{goyal-durrett-2020-evaluating, laban2022summac} and QA-based \citep{scialom-etal-2021-questeval, fabbri-etal-2022-qafacteval} approaches primarily focus on the faithfulness dimension. LLM-based evaluators \citep{liu-etal-2023-g, fu-etal-2024-gptscore} can easily assess multiple dimensions but suffer from instability and coarse granularity. Recently, several works \citep{song-etal-2024-finesure, wan-etal-2024-acueval, yang-etal-2024-fizz, scire-etal-2024-fenice, jeong-etal-2025-agent, koupaee-etal-2025-faithful} have achieved fine-grained and interpretable evaluation through atomic facts \citep{liu-etal-2023-revisiting, min-etal-2023-factscore} and LLM-based claim verification \citep{song-etal-2024-veriscore, tang-etal-2024-minicheck, oh-etal-2025-learning, chen-etal-2025-graphcheck}. Following this paradigm, we integrate complementary key fact alignment \citep{song-etal-2024-finesure} and summary fact verification \citep{wan-etal-2024-acueval} to deliver a fully fact-level evaluation across all dimensions.

\section{\textsc{OmniCSEval}}
\subsection{Conversation Collection}
\vspace{1pt}
\paragraph{Data Sources.}
We aggregate dialogues from 13 established datasets across six real-world scenarios. \textbf{Daily Life} captures informal exchanges from SAMSum \citep{gliwa-etal-2019-samsum} and DialogSum \citep{chen-etal-2021-dialogsum}. \textbf{Media Interview} utilizes MediaSum \citep{zhu-etal-2021-mediasum} for broadcast discourse. \textbf{Screenplay} incorporates SummScreen \citep{chen-etal-2022-summscreen} and MovieSum \citep{saxena-keller-2024-moviesum}, characterized by extensive narratives and rich character interactions. For \textbf{Customer Service}, TweetSumm \citep{feigenblat-etal-2021-tweetsumm-dialog} and TODSum \citep{zhao2021todsumtaskorienteddialoguesummarization} provide task-oriented, problem-solving conversations. \textbf{Healthcare} features specialized clinical consultations sourced from ACI-Bench \citep{yim2023acibenchnovelambientclinical}, PriMock57 \citep{papadopoulos-korfiatis-etal-2022-primock57}, and MTS-Dialog \citep{ben-abacha-etal-2023-empirical}. Finally, \textbf{Meeting} leverages MeetingBank \citep{hu-etal-2023-meetingbank}, QMSum \citep{zhong-etal-2021-qmsum}, and ECTSum \citep{mukherjee-etal-2022-ectsum} to represent multi-party professional discussions. Detailed dataset descriptions are provided in Appendix \ref{sec:dataset_details}.

\begin{table}[t]
    \centering
    \small
    \setlength{\abovecaptionskip}{8pt}
    \setlength{\tabcolsep}{5.0pt}
    \resizebox{\linewidth}{!}{
    \begin{tabular}{lcccc}
        \toprule[1pt]
        \textbf{Scenario} & \textbf{Samples} & \textbf{\#Tokens} & \textbf{\#Sents.} & \textbf{\#Spks.} \\
        \midrule[0.5pt]
        \rowcolor{gray!20}
        \multicolumn{5}{l}{\textit{Open-Ended}} \\
        Daily Life & 300 & 214.9 & 19.8 & 2.1 \\
        Media Interview & 300 & 1,682.5 & 90.9 & 5.8 \\
        Screenplay & 300 & 12,867.8 & 1,240.6 & 19.2 \\
        \midrule[0.5pt]
        \rowcolor{gray!20}
        \multicolumn{5}{l}{\textit{Private}} \\
        Customer Service & 300 & 266.4 & 22.0 & 2.0 \\
       	Healthcare & 300 & 1,073.6 & 78.4 & 2.0 \\
       	Meeting & 300 & 10,261.4 & 644.6 & 8.1 \\
        \midrule[0.5pt]
        \textbf{Total} & 1,800 & 4,394.4 & 349.4 & 6.5 \\
        \bottomrule[1pt]
    \end{tabular}
    }
    \caption{Conversation statistics across the six scenarios. \#Tokens, \#Sents., and \#Spks. denote the average number of tokens, sentences, and speakers, respectively.}
    \label{tab:benchmark_statistics}
    \vspace{-3pt}
\end{table}

\paragraph{Filtering and Sampling.} To ensure data quality, we first eliminate near-duplicates using a pairwise Jaccard similarity threshold of 0.70. Next, we apply a bilateral interquartile range filter (scale factor 1.5) to remove length outliers within each dataset, and then constrain conversation lengths to between 128 and 32k tokens\footnote{We use the \texttt{gpt-4o} tokenizer from \href{https://github.com/openai/tiktoken}{tiktoken}.}. Finally, we conduct stratified random sampling to draw 300 instances per scenario, establishing a final benchmark of 1,800 diverse evaluations. Detailed sampling quotas and statistics are provided in Table~\ref{tab:sampled_data_details} (Appendix~\ref{sec:sampled_data_details}).

\vspace{-1pt}
\paragraph{Data Analysis.}
As shown in Table~\ref{tab:benchmark_statistics}, the scenarios exhibit diverse intrinsic structural complexities, ranging from concise interactions in Customer Service and Daily Life to lengthy dialogues in Meeting and Screenplay. Beyond length, the conversational dynamics vary significantly, spanning from dyadic exchanges to multi-party environments, highlighting the comprehensiveness of our benchmark.

\subsection{Summary Generation}
We evaluate a comprehensive suite of 28 LLMs, categorized into four groups based on reasoning capability and model scale\footnote{Models with a \texttt{Thinking} suffix explicitly generate intermediate reasoning, whereas \texttt{Instruct} systems bypass explicit thinking processes.}. For the \textbf{Large Reasoning} models, we select GPT-5, Gemini 2.5 Pro, Qwen3-235B-A22B-Thinking, DeepSeek-R1, GLM-4.6-Thinking, Kimi-K2-Thinking, and MiniMax-M2. The \textbf{Small Reasoning} category consists of GPT-5-mini, GPT-5-nano, Gemini 2.5 Flash, Gemini 2.5 Flash Lite, Qwen3-14B-Thinking, Qwen3-8B-Thinking, and GLM-Z1-9B. Regarding \textbf{Large Instruction} systems, we evaluate GPT-4.1, GPT-4o, Qwen2.5-72B-Instruct-128K, Qwen3-235B-A22B-Instruct, DeepSeek-V3, and GLM-4.6-Instruct. Finally, the \textbf{Small Instruction} group comprises GPT-4.1-mini, GPT-4.1-nano, GPT-4o-mini, Gemini 2.0 Flash, Gemini 2.0 Flash Lite, Qwen3-14B-Instruct, Qwen3-8B-Instruct, and GLM-4-9B-Chat. See Table~\ref{tab:evaluated_llm_details} (Appendix~\ref{sec:evaluated_llm_details}) for detailed specifications, such as versions, sizes, and reasoning configurations.

To assess the intrinsic capabilities of LLMs, we apply a standardized zero-shot instruction: \textit{Summarize the above conversation}. To minimize randomness, we set the temperature to zero. In total, we obtain 50,400 summaries (1,800 samples $\times$ 28 models) for evaluation.

\begin{figure}[t]
  \centering
  \setlength{\abovecaptionskip}{8pt}
  \includegraphics[width=0.99\linewidth]{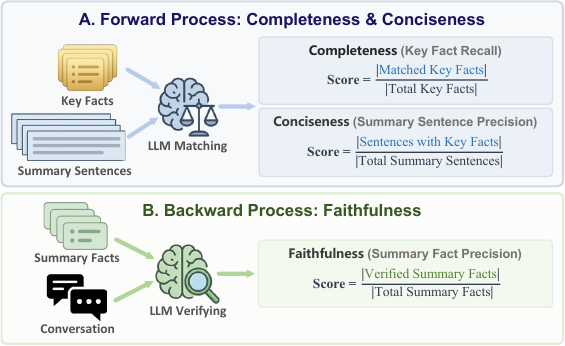}
  \caption{Overview of bidirectional fact-checking.}
  \label{fig:assessment_framework}
  \vspace{-3pt}
\end{figure}

\subsection{Annotation and Evaluation}
We focus on three key aspects of summary quality\footnote{We exclude \textit{fluency} and \textit{coherence} because current LLMs perform well on these dimensions \cite{song-etal-2025-learning}.}: \textbf{Completeness} (whether the summary captures all key content), \textbf{Conciseness} (whether the summary avoids non-salient information), and \textbf{Faithfulness} (whether the summary is factually consistent with the source dialogue). Figure~\ref{fig:assessment_framework} outlines the bidirectional fact-checking framework. It integrates key fact matching \citep{song-etal-2024-finesure} and summary fact verification \citep{wan-etal-2024-acueval} to perform an interpretable, fully fact-level assessment. Since the reliability of this framework hinges on the quality of evaluation units, we establish rigorous pipelines to curate key and summary facts.

\begin{table}[t]
    \centering
    \small
    \setlength{\abovecaptionskip}{8pt}
    \setlength{\tabcolsep}{6.0pt}
    \resizebox{\linewidth}{!}{
    \begin{tabular}{lcccc}
        \toprule[1pt]
        \textbf{Scenario} & \textbf{Self-Rej.} & \textbf{IAA} & \textbf{Agreed} & \textbf{Final} \\
        \midrule[0.5pt]
        \rowcolor{gray!20}
        \multicolumn{5}{l}{\textit{Open-Ended}} \\
        Daily Life & 547 & 0.612 & 219 & 2,572 \\
        Media Interview & 763 & 0.594 & 226 & 4,518 \\
        Screenplay & 796 & 0.566 & 339 & 5,292 \\
        \midrule[0.5pt]
        \rowcolor{gray!20}
        \multicolumn{5}{l}{\textit{Private}} \\
        Customer Service & 647 & 0.454 & 289 & 2,979 \\
        Healthcare & 1,039 & 0.672 & 231 & 6,414 \\
        Meeting & 967 & 0.595 & 355 & 5,856 \\
        \midrule[0.5pt]
        \textbf{Total} & 4,759 & 0.601 & 1,659 & 27,631 \\
        \bottomrule[1pt]
    \end{tabular}
    }
    \caption{Key fact annotation statistics. Self-Rej. counts rejected facts. IAA reports Cohen's $\kappa$. Agreed indicates the number of facts unanimously accepted as key, and Final represents the total number of key facts collected.}
    \label{tab:key_fact_annotation}
    \vspace{-3pt}
\end{table}

\vspace{-1pt}
\paragraph{Key and Summary Fact Collection.}
We annotate key facts using a three-stage human-LLM collaborative pipeline comprising preliminary extraction, self-review, and human adjudication. First, we utilize the advanced Gemini 3 Pro (high thinking-level) to extract candidate key facts from the source conversation adhering to importance and atomicity. The model then performs a second-round self-reflection to scrutinize these facts and filter out non-essential ones. Candidates consistently retained across both rounds are automatically accepted as high-confidence key facts, while those rejected during self-review undergo human adjudication.

To ensure annotation quality, we enlist four NLP experts (i.e., the first four authors). They are organized into two decision groups, with each member independently re-evaluating the rejected facts. To mitigate cognitive load, we also implement context mapping \citep{lee-etal-2024-unisumeval}, highlighting up to 25\scalebox{0.90}{$\%$} (capped at 2,048 tokens for lengthy dialogues) of the source sentences that are most semantically similar to the rejected facts\footnote{We use \texttt{text-embedding-3-small} from OpenAI for sentence and fact embeddings.}. To facilitate accurate judgment, we further provide annotators with the specific rejection reasons generated by Gemini 3 Pro. By synthesizing the focused context with the model's critique, the experts make a final binary determination (key or non-key) for each fact. As shown in Table~\ref{tab:key_fact_annotation}, this annotation achieves substantial inter-annotator agreement, yielding a Cohen's $\kappa$ of 0.60. To ensure criticality, only facts unanimously agreed upon as key are integrated into the final dataset. Detailed statistics and instructions are provided in Appendices~\ref{sec:key_fact_details} and \ref{sec:instructions}, respectively.

\begin{figure}[t]
  \centering
  \setlength{\abovecaptionskip}{5pt}
  \includegraphics[width=0.999\linewidth]{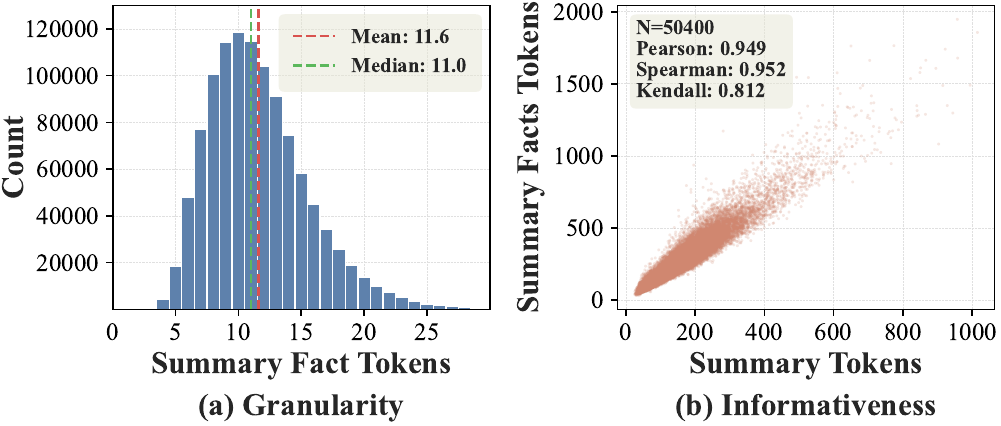}
  \caption{Statistics of collected summary facts.}
  \label{fig:combined_analysis_plot}
  \vspace{-5pt}
\end{figure}

For summary fact processing, we implement a two-stage pipeline comprising atomic decomposition and multi-LLM consensus filtering. Initially, we utilize DeepSeek-V3.2-Instruct to decompose all 50,400 summaries into atomic facts adhering to losslessness and atomicity. To prevent hallucinated artifacts from corrupting downstream faithfulness evaluations, we verify the entailment of each fact against its source summary through a majority vote among GLM-4.7-Instruct, Kimi-K2-Instruct, and Qwen3-Coder-480B-A35B. Facts deemed unsupported by at least two judges are classified as hallucinations and discarded. Notably, only 0.22\scalebox{0.90}{$\%$} (2,387) of the 1.1 million facts are flagged, underscoring the high reliability of our decomposition process. Figure~\ref{fig:combined_analysis_plot} further validates the fine granularity and losslessness of the decomposition. Detailed instructions are provided in Appendix~\ref{sec:instructions}.

\vspace{6pt}
\noindent\textbf{Bidirectional Fact-Checking and Scoring.} We leverage DeepSeek-V3.2-Instruct as the backbone model to execute this framework. Forward matching determines the semantic entailment of key facts by summary sentences, whereas backward verification identifies whether summary facts are supported by the source conversation. Consequently, completeness is measured as the recall of key facts, conciseness is quantified by the density of informative summary sentences, and faithfulness is calculated as the precision of trustworthy summary facts (see Appendices~\ref{sec:formula_details} and \ref{sec:instructions} for detailed formulas and instructions). In meta-evaluation (\S\ref{sec:meta_evaluation}), we show the stability of results, the robustness of leaderboards, and fact-level alignment with human judgments.

\section{Results and Analysis}
\subsection{Main Results}
\label{sec:main_results}
\begin{figure}[t]
  \centering
  \setlength{\abovecaptionskip}{5pt}
  \includegraphics[width=0.999\linewidth]{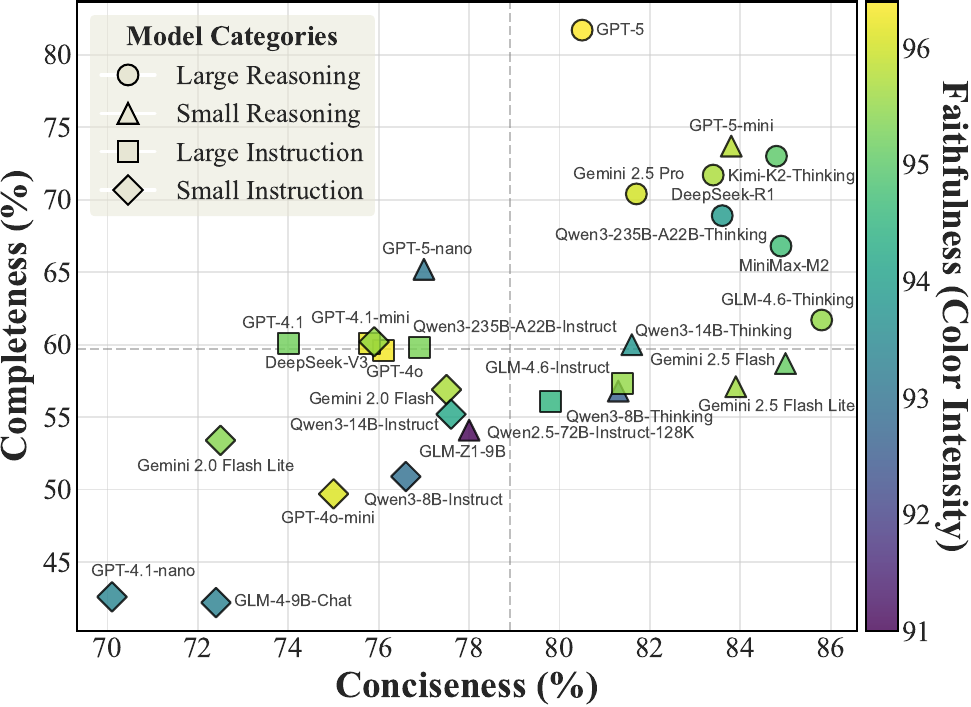}
  \caption{\textbf{Overall performance landscape}. Large reasoning models lead in both completeness and conciseness. Small reasoning models match large instruction models in completeness but outperform them in conciseness, whereas small instruction models lag behind.}
  \label{fig:main_results_figure1}
  \vspace{-5pt}
\end{figure}

\vspace{1pt}
Figure \ref{fig:main_results_figure1} and Table \ref{tab:main_evaluation_results} present the overall landscape and evaluation results across 28 LLMs. We highlight several critical findings regarding scenario challenges and the impact of explicit reasoning and model scale on conversation summarization.

\vspace{-2pt}
\paragraph{Cross-Scenario Challenges.}
As presented in Table~\ref{tab:main_evaluation_results}, no single scenario yields the highest scores across all dimensions, indicating that \textit{different scenarios present distinct difficulties}. In Meeting and Screenplay, LLMs exhibit a pronounced decline in both completeness and conciseness, with Meeting generally suffering the poorest conciseness. This highlights that current LLMs still struggle to accurately and effectively capture essential information from extensive and entangled conversations. Media Interview demonstrates similar, albeit less severe, challenges in these two dimensions. Intriguingly, while LLMs achieve respectable completeness and conciseness in the seemingly simple Daily Life scenario, they overwhelmingly exhibit the worst faithfulness. This underscores a severe vulnerability to hallucinations when processing highly open-ended and informal interactions. Furthermore, Screenplay and Customer Service also pose moderate-to-high faithfulness challenges. Finally, the moderate hallucination rates observed in Healthcare represent a critical concern, given the high stakes and strict factual requirements of medical documentation.

\definecolor{beanpaste}{RGB}{204, 232, 207}

\newcommand{\hl}[1]{{%
  \setlength{\fboxsep}{0.8pt}%
  \colorbox{beanpaste!70}{#1}%
}}

\newcommand{\pad}[1]{{%
  \setlength{\fboxsep}{0.8pt}%
  \setlength{\fboxrule}{0pt}%
  \fbox{#1}%
}}

\begin{table*}[t]
    \centering
    \small
    \setlength{\abovecaptionskip}{5pt}
    \setlength{\tabcolsep}{4.5pt}
    \resizebox{\textwidth}{!}{
    \begin{tabular}{lccccccc}
        \toprule[1pt]
        \multirow{4.2}{*}{\textbf{Model Name}} &
        \multicolumn{7}{c}{\textbf{\textsc{OmniCSEval}} (\textit{Multi-dimensional evaluation results for} \textbf{Completeness} / \textbf{Conciseness} / \textbf{Faithfulness})} \\
        \cmidrule(lr){2-8}
        & \multicolumn{3}{c}{\textit{Open-Ended}} & 
        \multicolumn{3}{c}{\textit{Private}} & 
        \multirow{2.5}{*}{\makecell{\textbf{Overall} \\ \textbf{Performance}}} \\
        \cmidrule(lr){2-4}
        \cmidrule(lr){5-7}
        & Screenplay & Interview & Daily & Meeting & Healthcare & Customer \\
        \midrule[0.5pt]
        \rowcolor{gray!20}
        \multicolumn{8}{l}{\textit{Large Reasoning Language Models}} \\
       	\rule{0pt}{10pt}GPT-5\,\raisebox{0.9pt}{{\tiny\textcolor{softred}{\faLock}}} & \multicolumn{1}{|c}{\hl{75.1} / \pad{76.7} / \hl{97.1}} & \hl{70.1} / \pad{81.3} / \hl{97.5} & \pad{81.8} / \pad{87.8} / \pad{93.1} & \multicolumn{1}{|c}{\hl{80.7} / \pad{63.6} / \hl{98.2}} & \hl{91.2} / \pad{90.3} / \hl{96.5} & \hl{91.1} / \pad{83.5} / \hl{95.8} & \multicolumn{1}{|c}{\hl{81.7} / \pad{80.5} / \hl{96.4}} \\
       	\rule{0pt}{8.65pt}Gemini 2.5 Pro\,\raisebox{0.9pt}{{\tiny\textcolor{softred}{\faLock}}} & \multicolumn{1}{|c}{\pad{63.1} / \pad{73.1} / \pad{96.3}} & \pad{60.1} / \pad{78.9} / \pad{96.4} & \pad{75.6} / \pad{90.2} / \pad{93.4} & \multicolumn{1}{|c}{\pad{61.6} / \pad{63.2} / \pad{97.7}} & \pad{80.2} / \pad{95.0} / \hl{96.5} & \pad{81.6} / \pad{89.6} / \pad{95.6} & \multicolumn{1}{|c}{\pad{70.4} / \pad{81.7} / \pad{96.0}} \\
        \rule{0pt}{8.65pt}Qwen3-235B-A22B-Thinking\,\raisebox{0.9pt}{{\tiny\textcolor{softgreen}{\faUnlock}}} & \multicolumn{1}{|c}{\pad{52.8} / \pad{74.4} / \pad{95.4}} & \pad{63.7} / \pad{83.9} / \pad{94.2} & \pad{79.0} / \hl{93.7} / \pad{90.2} & \multicolumn{1}{|c}{\pad{55.0} / \pad{71.5} / \pad{96.3}} & \pad{78.9} / \pad{92.2} / \pad{93.8} & \pad{84.1} / \pad{85.8} / \pad{93.3} & \multicolumn{1}{|c}{\pad{68.9} / \pad{83.6} / \pad{93.9}} \\
        \rule{0pt}{8.65pt}DeepSeek-R1\,\raisebox{0.9pt}{{\tiny\textcolor{softgreen}{\faUnlock}}} & \multicolumn{1}{|c}{\pad{57.8} / \pad{79.1} / \pad{96.4}} & \pad{62.2} / \pad{82.9} / \pad{95.8} & \hl{86.4} / \pad{87.0} / \hl{93.8} & \multicolumn{1}{|c}{\pad{53.9} / \hl{74.4} / \pad{97.0}} & \pad{81.6} / \pad{92.5} / \pad{95.6} & \pad{88.3} / \pad{84.5} / \pad{95.6} & \multicolumn{1}{|c}{\pad{71.7} / \pad{83.4} / \pad{95.7}} \\
        \rule{0pt}{8.65pt}GLM-4.6-Thinking\,\raisebox{0.9pt}{{\tiny\textcolor{softgreen}{\faUnlock}}} & \multicolumn{1}{|c}{\pad{51.2} / \pad{80.0} / \pad{95.4}} & \pad{56.3} / \pad{83.5} / \pad{97.2} & \pad{67.9} / \pad{91.0} / \pad{92.0} & \multicolumn{1}{|c}{\pad{51.0} / \pad{74.2} / \pad{97.3}} & \pad{67.9} / \hl{95.1} / \pad{96.4} & \pad{76.1} / \hl{90.8} / \pad{94.8} & \multicolumn{1}{|c}{\pad{61.7} / \hl{85.8} / \pad{95.5}} \\
        \rule{0pt}{8.65pt}Kimi-K2-Thinking\,\raisebox{0.9pt}{{\tiny\textcolor{softgreen}{\faUnlock}}} & \multicolumn{1}{|c}{\pad{63.0} / \pad{81.8} / \pad{95.6}} & \pad{65.3} / \pad{83.1} / \pad{95.8} & \pad{79.4} / \pad{90.2} / \pad{92.2} & \multicolumn{1}{|c}{\pad{62.8} / \pad{72.9} / \pad{96.8}} & \pad{82.3} / \pad{93.0} / \pad{95.8} & \pad{85.2} / \pad{88.1} / \pad{94.0} & \multicolumn{1}{|c}{\pad{73.0} / \pad{84.8} / \pad{95.0}} \\
        \rule{0pt}{8.65pt}MiniMax-M2\,\raisebox{0.9pt}{{\tiny\textcolor{softgreen}{\faUnlock}}} & \multicolumn{1}{|c}{\pad{51.3} / \hl{84.2} / \pad{93.8}} & \pad{56.8} / \hl{84.2} / \pad{95.8} & \pad{76.6} / \pad{90.2} / \pad{92.3} & \multicolumn{1}{|c}{\pad{57.1} / \pad{71.8} / \pad{96.7}} & \pad{75.4} / \pad{92.7} / \pad{94.9} & \pad{83.2} / \pad{86.3} / \pad{95.0} & \multicolumn{1}{|c}{\pad{66.8} / \pad{84.9} / \pad{94.7}} \\
        \midrule[0.5pt]
        \rowcolor{gray!20}
        \multicolumn{8}{l}{\textit{Small Reasoning Language Models}} \\
       	\rule{0pt}{10pt}GPT-5-mini\,\raisebox{0.9pt}{{\tiny\textcolor{softred}{\faLock}}} & \multicolumn{1}{|c}{\pad{51.8} / \pad{77.9} / \hl{95.7}} & \hl{63.7} / \hl{87.0} / \pad{96.4} & \hl{81.4} / \hl{92.3} / \pad{92.6} & \multicolumn{1}{|c}{\hl{67.9} / \pad{70.5} / \pad{97.3}} & \hl{85.8} / \pad{93.0} / \hl{96.4} & \hl{91.5} / \pad{82.1} / \hl{96.2} & \multicolumn{1}{|c}{\hl{73.7} / \pad{83.8} / \hl{95.8}} \\
       	\rule{0pt}{8.65pt}GPT-5-nano\,\raisebox{0.9pt}{{\tiny\textcolor{softred}{\faLock}}} & \multicolumn{1}{|c}{\pad{34.5} / \pad{66.6} / \pad{90.9}} & \pad{58.5} / \pad{79.8} / \pad{94.1} & \pad{75.7} / \pad{86.1} / \pad{89.4} & \multicolumn{1}{|c}{\pad{54.7} / \pad{64.1} / \pad{95.7}} & \pad{80.1} / \pad{87.5} / \pad{94.2} & \pad{87.4} / \pad{78.1} / \pad{94.5} & \multicolumn{1}{|c}{\pad{65.2} / \pad{77.0} / \pad{93.1}} \\
       	\rule{0pt}{8.65pt}Gemini 2.5 Flash\,\raisebox{0.9pt}{{\tiny\textcolor{softred}{\faLock}}} & \multicolumn{1}{|c}{\hl{56.5} / \hl{82.1} / \pad{95.3}} & \pad{48.6} / \pad{81.3} / \pad{95.9} & \pad{67.0} / \pad{90.2} / \pad{91.9} & \multicolumn{1}{|c}{\pad{46.2} / \hl{70.8} / \pad{97.1}} & \pad{65.7} / \hl{95.7} / \pad{96.0} & \pad{68.5} / \pad{89.5} / \pad{94.4} & \multicolumn{1}{|c}{\pad{58.7} / \hl{85.0} / \pad{95.1}} \\
       	\rule{0pt}{8.65pt}Gemini 2.5 Flash Lite\,\raisebox{0.9pt}{{\tiny\textcolor{softred}{\faLock}}} & \multicolumn{1}{|c}{\pad{44.4} / \pad{75.0} / \pad{94.8}} & \pad{51.1} / \pad{82.3} / \hl{97.3} & \pad{66.2} / \pad{90.6} / \hl{92.7} & \multicolumn{1}{|c}{\pad{44.8} / \pad{68.8} / \hl{97.5}} & \pad{66.6} / \pad{93.6} / \hl{96.4} & \pad{69.8} / \hl{92.9} / \pad{95.1} & \multicolumn{1}{|c}{\pad{57.1} / \pad{83.9} / \pad{95.6}} \\
        \rule{0pt}{8.65pt}Qwen3-14B-Thinking\,\raisebox{0.9pt}{{\tiny\textcolor{softgreen}{\faUnlock}}} & \multicolumn{1}{|c}{\pad{36.5} / \pad{69.4} / \pad{91.8}} & \pad{53.0} / \pad{85.1} / \pad{95.2} & \pad{76.4} / \pad{89.2} / \pad{90.6} & \multicolumn{1}{|c}{\pad{43.5} / \pad{69.9} / \pad{95.5}} & \pad{70.0} / \pad{92.6} / \pad{95.0} & \pad{80.7} / \pad{83.2} / \pad{95.0} & \multicolumn{1}{|c}{\pad{60.0} / \pad{81.6} / \pad{93.8}} \\
        \rule{0pt}{8.65pt}Qwen3-8B-Thinking\,\raisebox{0.9pt}{{\tiny\textcolor{softgreen}{\faUnlock}}} & \multicolumn{1}{|c}{\pad{31.6} / \pad{64.7} / \pad{89.5}} & \pad{52.0} / \pad{85.3} / \pad{93.7} & \pad{70.6} / \pad{90.0} / \pad{89.9} & \multicolumn{1}{|c}{\pad{42.3} / \pad{69.9} / \pad{94.3}} & \pad{67.5} / \pad{93.6} / \pad{94.5} & \pad{76.9} / \pad{84.1} / \pad{93.7} & \multicolumn{1}{|c}{\pad{56.8} / \pad{81.3} / \pad{92.6}} \\
        \rule{0pt}{8.65pt}GLM-Z1-9B\,\raisebox{0.9pt}{{\tiny\textcolor{softgreen}{\faUnlock}}} & \multicolumn{1}{|c}{\pad{26.5} / \pad{56.9} / \pad{85.7}} & \pad{50.5} / \pad{81.5} / \pad{92.9} & \pad{68.9} / \pad{87.8} / \pad{88.7} & \multicolumn{1}{|c}{\pad{39.3} / \pad{67.9} / \pad{92.6}} & \pad{63.4} / \pad{91.3} / \pad{93.1} & \pad{75.9} / \pad{82.5} / \pad{93.1} & \multicolumn{1}{|c}{\pad{54.1} / \pad{78.0} / \pad{91.0}} \\
        \midrule[0.5pt]
        \rowcolor{gray!20}
        \multicolumn{8}{l}{\textit{Large Instruction Language Models}} \\
        \rule{0pt}{10pt}GPT-4.1\,\raisebox{0.9pt}{{\tiny\textcolor{softred}{\faLock}}} & \multicolumn{1}{|c}{\hl{44.6} / \pad{63.5} / \pad{96.0}} & \pad{48.2} / \pad{71.1} / \pad{95.9} & \pad{71.5} / \pad{84.8} / \pad{92.0} & \multicolumn{1}{|c}{\hl{42.6} / \pad{54.6} / \pad{97.3}} & \hl{76.3} / \pad{89.0} / \pad{95.5} & \pad{77.4} / \pad{80.9} / \pad{94.7} & \multicolumn{1}{|c}{\hl{60.1} / \pad{74.0} / \pad{95.2}} \\
        \rule{0pt}{8.65pt}GPT-4o\,\raisebox{0.9pt}{{\tiny\textcolor{softred}{\faLock}}} & \multicolumn{1}{|c}{\pad{42.5} / \pad{66.4} / \hl{96.8}} & \pad{50.4} / \pad{74.0} / \hl{96.7} & \pad{74.1} / \pad{87.7} / \hl{93.9} & \multicolumn{1}{|c}{\pad{36.5} / \pad{56.0} / \pad{97.6}} & \pad{73.7} / \pad{90.3} / \hl{96.5} & \pad{80.2} / \pad{82.2} / \pad{96.1} & \multicolumn{1}{|c}{\pad{59.6} / \pad{76.1} / \hl{96.3}} \\
        \rule{0pt}{8.65pt}Qwen3-235B-A22B-Instruct\,\raisebox{0.9pt}{{\tiny\textcolor{softgreen}{\faUnlock}}} & \multicolumn{1}{|c}{\pad{42.9} / \pad{65.5} / \pad{95.6}} & \hl{51.3} / \pad{76.6} / \pad{96.4} & \pad{74.1} / \pad{85.0} / \pad{91.7} & \multicolumn{1}{|c}{\pad{41.9} / \pad{62.3} / \pad{96.7}} & \pad{72.3} / \pad{90.8} / \pad{95.8} & \pad{76.4} / \pad{81.1} / \pad{95.7} & \multicolumn{1}{|c}{\pad{59.8} / \pad{76.9} / \pad{95.3}} \\
        \rule{0pt}{8.65pt}Qwen2.5-72B-Instruct-128K\,\raisebox{0.9pt}{{\tiny\textcolor{softgreen}{\faUnlock}}} & \multicolumn{1}{|c}{\pad{39.2} / \pad{71.8} / \pad{94.7}} & \pad{47.5} / \hl{82.6} / \pad{95.6} & \pad{72.3} / \hl{87.9} / \pad{91.9} & \multicolumn{1}{|c}{\pad{35.6} / \pad{62.6} / \pad{95.8}} & \pad{64.9} / \pad{91.6} / \pad{94.6} & \pad{77.1} / \pad{82.1} / \pad{94.4} & \multicolumn{1}{|c}{\pad{56.1} / \pad{79.8} / \pad{94.5}} \\
        \rule{0pt}{8.65pt}DeepSeek-V3\,\raisebox{0.9pt}{{\tiny\textcolor{softgreen}{\faUnlock}}} & \multicolumn{1}{|c}{\pad{39.0} / \pad{65.2} / \pad{96.7}} & \pad{50.8} / \pad{75.9} / \pad{96.6} & \hl{82.1} / \pad{85.7} / \pad{93.5} & \multicolumn{1}{|c}{\pad{34.2} / \pad{58.7} / \hl{97.9}} & \pad{71.8} / \pad{91.0} / \pad{96.2} & \hl{82.5} / \pad{78.3} / \hl{96.3} & \multicolumn{1}{|c}{\hl{60.1} / \pad{75.8} / \pad{96.2}} \\
        \rule{0pt}{8.65pt}GLM-4.6-Instruct\,\raisebox{0.9pt}{{\tiny\textcolor{softgreen}{\faUnlock}}} & \multicolumn{1}{|c}{\hl{44.6} / \hl{75.0} / \pad{95.5}} & \pad{47.0} / \pad{80.8} / \pad{95.9} & \pad{72.6} / \pad{87.6} / \pad{92.8} & \multicolumn{1}{|c}{\pad{38.3} / \hl{67.2} / \pad{97.8}} & \pad{64.8} / \hl{93.4} / \pad{95.6} & \pad{76.4} / \hl{84.2} / \pad{95.2} & \multicolumn{1}{|c}{\pad{57.3} / \hl{81.4} / \pad{95.5}} \\
        \midrule[0.5pt]
        \rowcolor{gray!20}
        \multicolumn{8}{l}{\textit{Small Instruction Language Models}} \\
        \rule{0pt}{10pt}GPT-4.1-mini\,\raisebox{0.9pt}{{\tiny\textcolor{softred}{\faLock}}} & \multicolumn{1}{|c}{\pad{36.4} / \pad{60.7} / \pad{95.2}} & \hl{51.8} / \pad{76.1} / \pad{96.0} & \hl{73.5} / \pad{87.2} / \pad{92.9} & \multicolumn{1}{|c}{\hl{46.1} / \pad{58.8} / \pad{97.2}} & \hl{72.6} / \pad{91.8} / \pad{95.9} & \hl{80.5} / \pad{80.5} / \hl{95.7} & \multicolumn{1}{|c}{\hl{60.2} / \pad{75.9} / \pad{95.5}} \\
        \rule{0pt}{8.65pt}GPT-4.1-nano\,\raisebox{0.9pt}{{\tiny\textcolor{softred}{\faLock}}} & \multicolumn{1}{|c}{\pad{16.8} / \pad{43.5} / \pad{93.5}} & \pad{38.5} / \pad{76.1} / \pad{94.1} & \pad{56.9} / \pad{83.8} / \pad{89.3} & \multicolumn{1}{|c}{\pad{20.9} / \pad{48.2} / \pad{96.6}} & \pad{58.1} / \pad{90.9} / \pad{94.0} & \pad{64.1} / \pad{77.9} / \pad{92.6} & \multicolumn{1}{|c}{\pad{42.6} / \pad{70.1} / \pad{93.4}} \\
        \rule{0pt}{8.65pt}GPT-4o-mini\,\raisebox{0.9pt}{{\tiny\textcolor{softred}{\faLock}}} & \multicolumn{1}{|c}{\pad{26.3} / \pad{56.5} / \hl{96.3}} & \pad{43.4} / \pad{77.3} / \hl{97.3} & \pad{68.9} / \hl{88.3} / \hl{93.5} & \multicolumn{1}{|c}{\pad{27.6} / \pad{55.8} / \pad{97.3}} & \pad{60.8} / \pad{91.1} / \pad{96.4} & \pad{71.2} / \pad{81.0} / \pad{95.6} & \multicolumn{1}{|c}{\pad{49.7} / \pad{75.0} / \hl{96.1}} \\
       	\rule{0pt}{8.65pt}Gemini 2.0 Flash\,\raisebox{0.9pt}{{\tiny\textcolor{softred}{\faLock}}} & \multicolumn{1}{|c}{\hl{43.0} / \hl{69.8} / \pad{95.0}} & \pad{46.5} / \pad{75.1} / \pad{96.6} & \pad{70.4} / \pad{88.2} / \pad{92.8} & \multicolumn{1}{|c}{\pad{37.2} / \pad{52.7} / \hl{98.0}} & \pad{70.5} / \pad{92.1} / \hl{96.5} & \pad{73.8} / \hl{86.9} / \pad{95.4} & \multicolumn{1}{|c}{\pad{56.9} / \pad{77.5} / \pad{95.7}} \\
       	\rule{0pt}{8.65pt}Gemini 2.0 Flash Lite\,\raisebox{0.9pt}{{\tiny\textcolor{softred}{\faLock}}} & \multicolumn{1}{|c}{\pad{32.0} / \pad{57.7} / \pad{95.0}} & \pad{46.1} / \pad{70.9} / \pad{97.0} & \pad{69.2} / \pad{83.2} / \pad{91.9} & \multicolumn{1}{|c}{\pad{34.2} / \pad{53.6} / \pad{97.8}} & \pad{63.9} / \pad{88.8} / \pad{96.0} & \pad{74.9} / \pad{80.6} / \pad{94.5} & \multicolumn{1}{|c}{\pad{53.4} / \pad{72.5} / \pad{95.4}} \\
        \rule{0pt}{8.65pt}Qwen3-14B-Instruct\,\raisebox{0.9pt}{{\tiny\textcolor{softgreen}{\faUnlock}}} & \multicolumn{1}{|c}{\pad{33.6} / \pad{62.9} / \pad{91.8}} & \pad{45.7} / \hl{81.8} / \pad{95.2} & \pad{73.1} / \pad{85.1} / \pad{90.7} & \multicolumn{1}{|c}{\pad{35.2} / \hl{62.3} / \pad{96.6}} & \pad{65.3} / \pad{92.0} / \pad{95.4} & \pad{78.2} / \pad{81.6} / \pad{95.5} & \multicolumn{1}{|c}{\pad{55.2} / \hl{77.6} / \pad{94.2}} \\
        \rule{0pt}{8.65pt}Qwen3-8B-Instruct\,\raisebox{0.9pt}{{\tiny\textcolor{softgreen}{\faUnlock}}} & \multicolumn{1}{|c}{\pad{29.3} / \pad{56.6} / \pad{89.5}} & \pad{43.0} / \pad{79.6} / \pad{94.3} & \pad{66.6} / \pad{87.8} / \pad{90.2} & \multicolumn{1}{|c}{\pad{34.7} / \pad{59.7} / \pad{94.9}} & \pad{61.6} / \hl{92.5} / \pad{94.2} & \pad{70.0} / \pad{83.4} / \pad{94.5} & \multicolumn{1}{|c}{\pad{50.9} / \pad{76.6} / \pad{92.9}} \\
        \rule{0pt}{8.65pt}GLM-4-9B-Chat\,\raisebox{0.9pt}{{\tiny\textcolor{softgreen}{\faUnlock}}} & \multicolumn{1}{|c}{\pad{20.1} / \pad{47.3} / \pad{91.2}} & \pad{36.7} / \pad{76.0} / \pad{95.1} & \pad{61.8} / \pad{87.2} / \pad{89.5} & \multicolumn{1}{|c}{\pad{19.2} / \pad{49.9} / \pad{96.9}} & \pad{50.7} / \pad{92.1} / \pad{94.1} & \pad{64.7} / \pad{82.1} / \pad{93.1} & \multicolumn{1}{|c}{\pad{42.2} / \pad{72.4} / \pad{93.3}} \\
        \bottomrule[1pt]
    \end{tabular}
    }
    \caption{Main evaluation results. \raisebox{0.5pt}{{\tiny\textcolor{softred}{\faLock}}} and \raisebox{0.5pt}{{\tiny\textcolor{softgreen}{\faUnlock}}} denote \textit{proprietary} and \textit{open-weights} systems, respectively. The values in each cell represent \textbf{Completeness}, \textbf{Conciseness}, and \textbf{Faithfulness} of the model-generated summaries, respectively. All reported scores are \textit{macro-averaged} (\scalebox{0.90}{$\%$}). We highlight the \textit{best} model for each dimension in each LLM category.}
    \label{tab:main_evaluation_results}
    \vspace{-8pt}
\end{table*}

\begin{figure*}[t]
  \centering
  \setlength{\abovecaptionskip}{0pt}
  \includegraphics[width=0.999\linewidth]{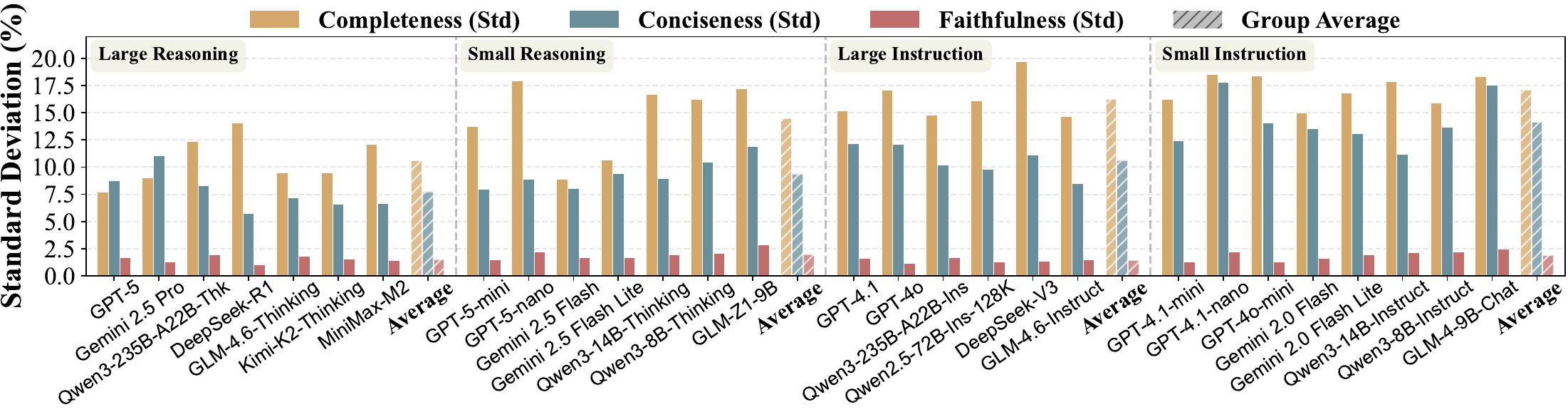}
  \caption{Stability of LLMs across six scenarios. Lower standard deviation indicates better performance consistency.}
  \label{fig:stability_analysis}
  \vspace{-5pt}
\end{figure*}

\vspace{-2pt}
\paragraph{Reasoning Is the Key to Completeness and Conciseness, but Not Faithfulness.}
As shown in Figure~\ref{fig:main_results_figure1} and Table~\ref{tab:main_evaluation_results}, reasoning systems demonstrate a decisive advantage in completeness and conciseness over instruction models of similar scale. However, they yield no gain, or even regress, in faithfulness. For instance, DeepSeek-R1 significantly outperforms DeepSeek-V3 in completeness (71.7\scalebox{0.90}{$\%$} vs. 60.1\scalebox{0.90}{$\%$}) and conciseness (83.4\scalebox{0.90}{$\%$} vs. 75.8\scalebox{0.90}{$\%$}), yet lags in faithfulness (95.7\scalebox{0.90}{$\%$} vs. 96.2\scalebox{0.90}{$\%$}, $p<0.01$). This trade-off is mirrored by Qwen3-235B-A22B, whose Thinking variant suffers a notable faithfulness drop compared to its Instruct version (95.3\scalebox{0.90}{$\%$} $\to$ 93.9\scalebox{0.90}{$\%$}). This tendency is pervasive across the benchmark: Kimi-K2-Thinking and MiniMax-M2 trail the majority of large instruction systems in faithfulness; GLM-4.6-Thinking merely ties its instruction equivalent (95.5\scalebox{0.90}{$\%$}); and even the SOTA GPT-5 exhibits no meaningful improvement over GPT-4o (96.4\scalebox{0.90}{$\%$} vs. 96.3\scalebox{0.90}{$\%$}, $p>0.05$). Smaller models, such as the Qwen3-8B and -14B series, echo this exact dynamic. These results reveal that \textit{while explicit reasoning is highly effective for key content identification and noise data filtering, it paradoxically increases the risk of hallucinations}. Furthermore, we find that the advantages of reasoning models in completeness and conciseness stem primarily from their robust performance in complex-context scenarios (i.e., Screenplay and Meeting), granting them superior cross-scenario stability in these two dimensions compared to similarly sized instruction models (see Figure~\ref{fig:stability_analysis}).

\noindent\textbf{Reasoning Offsets Scale, but Size Still Matters.}
As shown in Figure~\ref{fig:main_results_figure1}, small reasoning models effectively \textit{punch above their weight}, matching the completeness and surpassing the conciseness of large instruction systems. This indicates that explicit reasoning can compensate for parameter deficiencies to a remarkable extent. However, scaling laws remain a fundamental constraint: larger models widely achieve superior multi-dimensional performance, especially in completeness and faithfulness, within both reasoning and instruction settings. Crucially, scale dictates robustness. As evidenced by Figure~\ref{fig:stability_analysis}, larger models maintain greater cross-scenario stability in both reasoning and instruction paradigms. These findings suggest that \textit{while reasoning serves as a powerful equalizer, model scale remains an indispensable prerequisite for comprehensive and robust conversation summarization}.

\begin{figure*}[t]
  \centering
  \setlength{\abovecaptionskip}{6pt}
  \includegraphics[width=0.999\linewidth]{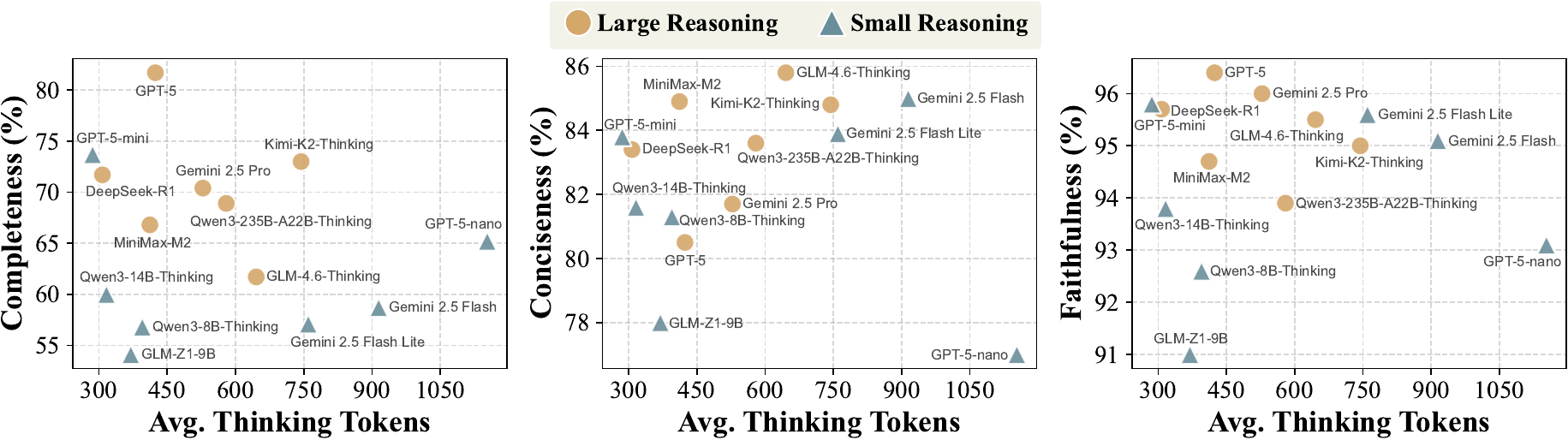}
  \caption{Visualization of average thinking tokens versus multi-dimensional performance. Models positioned closer to the upper-left corner demonstrate superior thinking efficiency, achieving high scores with minimal reasoning cost.}
  \label{fig:reasoning_efficiency}
  \vspace{-8pt}
\end{figure*}

\begin{figure}[t]
  \centering
  \setlength{\abovecaptionskip}{3pt}
  \includegraphics[width=0.999\linewidth]{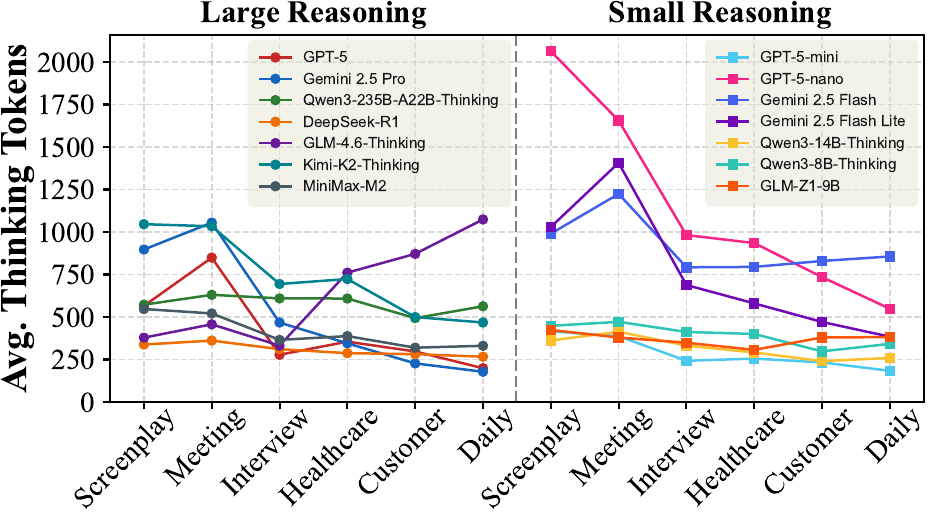}
  \caption{Thinking length distribution of reasoning models across scenarios, sorted by context structural complexity (i.e., dialogue length and number of speakers).}
  \label{fig:scenario_thinking_tokens}
  \vspace{-3pt}
\end{figure}

\vspace{-3pt}
\subsection{Further Analysis}
\paragraph{Thinking Efficiency of Reasoning Models.}
As Figure~\ref{fig:reasoning_efficiency} shows, larger reasoning models densely cluster in the upper-left region, demonstrating superior and more consistent thinking efficiency. Crucially, models that think longer do not necessarily achieve greater performance, particularly in completeness and faithfulness. For instance, Gemini 2.5 Flash and GLM-4.6-Thinking consume significantly more tokens than highly efficient peers such as GPT-5-mini and GPT-5, yet fail to achieve comparable completeness and faithfulness. A conspicuous outlier is GPT-5-nano, which exhibits the highest token consumption yet remains in the middle-to-lower performance tier for conciseness and faithfulness. These results underscore that \textit{thinking efficiently is far more critical than simply thinking extensively for conversation summarization}.

\paragraph{Reasoning Adaptability to Scenario Complexity.}
Beyond efficiency, a highly adaptable reasoning model should dynamically scale its reasoning budget to match scenario structural complexity. As Figure~\ref{fig:scenario_thinking_tokens} illustrates, \textit{models exhibit distinctly different strategies for allocating thinking resources}. Flexible thinkers (e.g., Kimi-K2-Thinking) demonstrate strong context awareness, systematically reducing their thinking duration as input complexity decreases. Conversely, rigid models (e.g., Qwen3-235B-A22B-Thinking and DeepSeek-R1) maintain a nearly constant reasoning budget across all scenarios. Most notably, GLM-4.6-Thinking displays an anomalous inverse trend, expending significantly more thinking tokens on lower-complexity scenarios. This highlights a critical misalignment in cognitive load allocation, suggesting the model struggles to accurately gauge task complexity. A similar inverse pattern appears among small models, where systems like GLM-Z1-9B and Gemini 2.5 Flash devote disproportionately heavy reasoning efforts to straightforward scenarios like Customer Service and Daily Life.

\begin{table}[t]
    \vspace{4pt}
    \centering
    \small
    \setlength{\abovecaptionskip}{8pt}
    \setlength{\tabcolsep}{5.5pt}
    \begin{tabular}{lcc}
        \toprule[1pt]
        \textbf{Dimension} & \textbf{System-Level} & \textbf{Summary-Level} \\
        \midrule[0.5pt]
        Completeness & 0.999 & 0.982 \textcolor{green}{${\uparrow 0.129}$} \\
        Conciseness & 0.995 & 0.962 \textcolor{green}{${\uparrow 0.054}$} \\
        Faithfulness & 0.977 & 0.912$^{*}$ \\
        \bottomrule[1pt]
    \end{tabular}
    \caption{Evaluation stability across three runs. System-level is quantified by Kendall's $W$ for ranking consistency, and summary-level by Krippendorff's $\alpha$ for score agreement. \textcolor{green}{${\uparrow}$} indicates the advantage of our implementation over FineSure \cite{song-etal-2024-finesure}. $^{*}$ represents no comparable results in ACUEval \citep{wan-etal-2024-acueval}.}
    \label{tab:stability_of_results}
    \vspace{-3pt}
\end{table}

\section{Meta Evaluation}
\label{sec:meta_evaluation}
\paragraph{Stability of Benchmark Results.}
Considering the inherent randomness of LLM evaluators (even with the temperature set to zero \cite{atil2025nondeterminismdeterministicllmsettings}), we assess the stability of our benchmark results to ensure reproducibility. For all 28 evaluated LLMs, we sample 300 summaries per model (50 per scenario). To guarantee representative coverage, this sampling is stratified by conversation length within each scenario, encompassing all sub-datasets and resulting in a total of 8,400 summaries. We then execute the bidirectional fact-checking framework three independent times using the DeepSeek-V3.2-Instruct backbone. As shown in Table \ref{tab:stability_of_results}, the evaluations exhibit exceptional consistency, confirming our results are highly stable and reproducible.

\noindent\textbf{Alignment with Human Judgments.} \quad To validate evaluation accuracy, we construct a human-annotated test set comprising 120 summaries (20 per scenario from 20 randomly selected LLMs), excluding the baseline fact-checker GPT-4o. We enlist six graduate students, divided into two groups, to perform key fact matching and summary fact verification, respectively. For the matching task, any complete initial disagreement among the three annotators triggers a secondary review until a majority consensus is reached. Finally, we determine the gold labels via majority voting, yielding a total of 1,871 matching and 2,625 verification annotations. Details of the annotation process and inter-annotator agreement are provided in Appendix~\ref{sec:human_annotation_details}.

We evaluate the performance of seven advanced LLMs as fact-checkers, ensuring that all evaluator models are strictly disjoint from the LLMs being evaluated. We employ Macro F1 \citep{magomere-etal-2026-distill} to assess the matching task and Balanced Accuracy (BAcc) \citep{tang-etal-2024-minicheck} for the verification task. As presented in Table~\ref{tab:pfc}, DeepSeek-V3.2-Instruct demonstrates exceptionally strong alignment with human judgments across both tasks, firmly indicating the accuracy of our evaluation.

\begin{table}[t]
    \centering
    \small
    \setlength{\abovecaptionskip}{5pt}
    \setlength{\tabcolsep}{3pt}
    \begin{tabular}{lcc}
        \toprule[1pt]
        \multirow{2.4}{*}{\textbf{LLM as a Fact-Checker}} & \textbf{Matching} & \textbf{Verification} \\[-1pt]
        \cmidrule(lr){2-2} \cmidrule(lr){3-3}\addlinespace[-1pt]
        & Macro F1 & BAcc \\
        \midrule[0.5pt]
        GPT-4o (\textit{Previous SOTA Judge}) & 0.935 & 0.774 \\
        \midrule[0.5pt]
        Qwen3-Next-80B-A3B-Instruct & 0.929 & 0.749 \\
        Llama 4 Maverick & 0.931 & 0.755 \\
        Mistral Large 3 & 0.931 & 0.762 \\
        Kimi-K2-Instruct & 0.953 & 0.805 \\
        GLM-4.7-Instruct & 0.946 & 0.853 \\
        DeepSeek-V3.2-Instruct & \textbf{0.958} & \textbf{0.871} \\
        \bottomrule[1pt]
    \end{tabular}
    \caption{Performance of various LLMs as fact-checkers.}
    \label{tab:pfc}
    \vspace{-5pt}
\end{table}

\begin{figure}[t]
  \centering
  \setlength{\abovecaptionskip}{3pt}
  \includegraphics[width=0.999\linewidth]{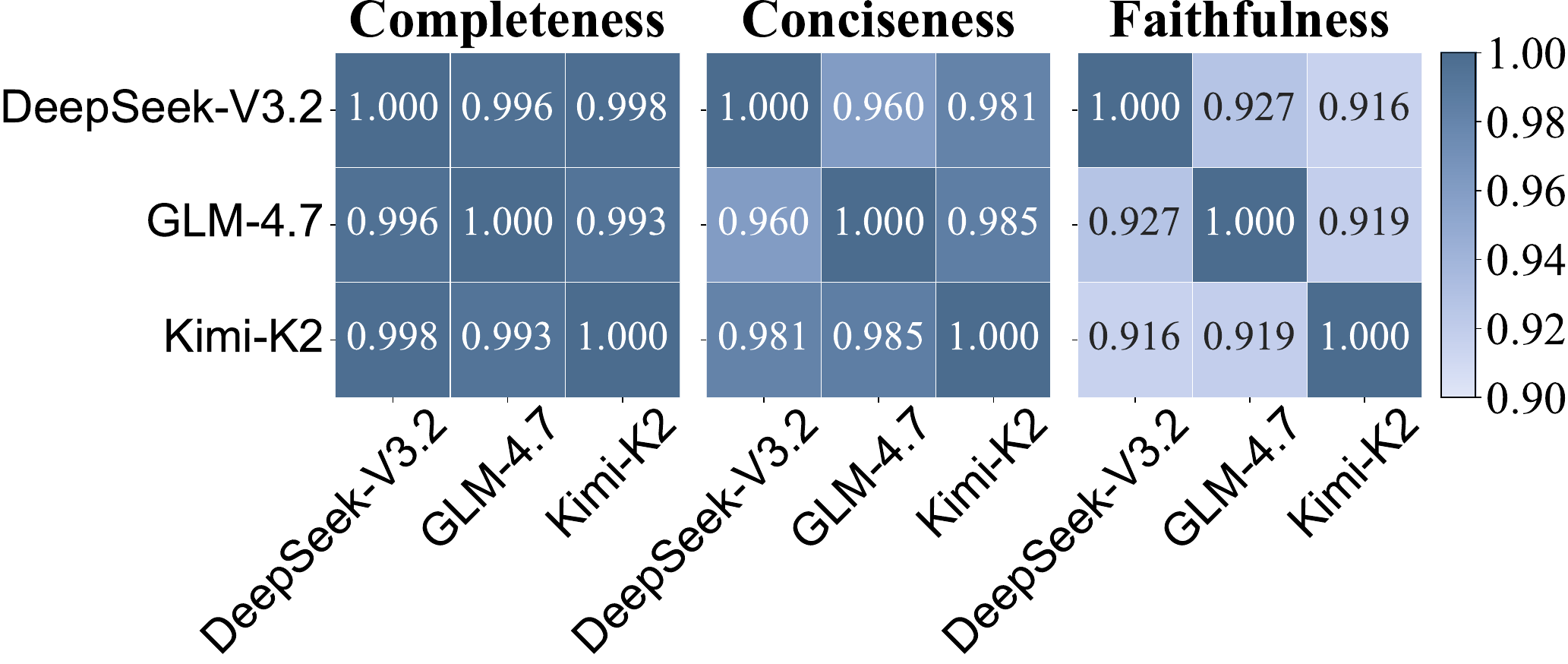}
  \caption{Pairwise system-level rank correlations among three distinct LLM judges (Spearman's $\rho$, all $p < 0.01$).}
  \label{fig:robustness_heatmap}
  \vspace{-5pt}
\end{figure}

\vspace{-1pt}
\paragraph{Leaderboard Robustness.}
A reliable benchmark leaderboard should be resilient to the choice of the automated LLM judge. To verify this, we re-evaluated the 8,400 summaries from our stability assessment (encompassing all 28 LLMs) using two additional advanced LLMs as backbones: GLM-4.7-Instruct and Kimi-K2-Instruct. As presented in Figure~\ref{fig:robustness_heatmap}, our evaluation demonstrates exceptional cross-judge consensus. Evaluators achieve near-perfect alignment in the highly discriminative completeness and conciseness (system-level average $\sigma=0.095$ and $0.049$, respectively). Even for faithfulness, which exhibits very low variance (average $\sigma=0.012$), rank correlations remain strong. These results confirm that the \textsc{OmniCSEval} leaderboard possesses robust, judge-agnostic reliability.

\section{Discussion}
\textbf{Which Model Should We Choose for Real-World Deployment?}
For applications demanding superior summarization quality, particularly within complex long-context scenarios, large reasoning models represent the optimal choice. Under budgetary constraints, small reasoning models serve as a powerful alternative, capable of rivaling large instruction models. Furthermore, reasoning systems are indispensable where concise summarization is prioritized. Conversely, for risk-averse tasks requiring hallucination mitigation, large instruction models remain the safest option. Finally, in resource-limited environments demanding low latency, small instruction models constitute the sole viable solution, assuming the quality trade-off is permissible.

\paragraph{Can Open-Weights Models Replace Proprietary Systems?}
Open-weights models now rival proprietary systems in large-scale reasoning and instruction. They often trade leads in completeness and conciseness. Despite this, a significant gap persists in small reasoning models where open-weights options lag behind SOTA leaders such as GPT-5-mini. Proprietary models also maintain superior faithfulness across all categories. This reliability makes them the safer choice for high-stakes environments.

\section{Conclusion}
In this study, we introduced \textsc{OmniCSEval}, a comprehensive and fine-grained benchmark designed to systematically evaluate LLMs in conversation summarization. Our empirical analysis reveals that LLMs face distinct, scenario-specific challenges, and demonstrates the impacts of explicit reasoning and model scale on multi-dimensional performance and cross-scenario stability. We also highlight the importance of reasoning efficiency and adaptability. We hope this evaluation work provides actionable guidance for real-world system deployment and establishes a rigorous foundation for future advancements in conversation summarization.

\section*{Limitations}
While \textsc{OmniCSEval} represents a significant advancement in the breadth and depth of conversation summarization evaluation, the current study focuses primarily on the English language. This constraint is intentionally adopted to prioritize the precision and reliability of the evaluation framework, thereby minimizing potential bias interference arising from cross-lingual semantic alignment issues. Recognizing that multilingual and cross-lingual capabilities are essential for general-purpose LLMs, we plan to explore broader linguistic environments in future dialogue summarization evaluation work.

\section*{Ethics Statement}
\textsc{OmniCSEval} aggregates 13 existing and publicly available datasets. While our benchmark includes domains that traditionally handle sensitive interactions, such as healthcare and customer service, we strictly utilized established corpora that have been thoroughly de-identified. These datasets were stripped of all personally identifiable information by their original creators. Consequently, this study does not introduce any new private or sensitive data into the public domain. For the meta-evaluation, we enlisted six graduate students. Prior to their participation, all annotators were fully informed of the study's purpose and provided explicit consent. To ensure fair labor practices, annotators were compensated at an hourly rate of \$8, which exceeds the local minimum wage requirements.

We acknowledge the substantial computational resources and environmental footprint required to conduct this study. Evaluating 28 LLMs to generate 50,400 summaries and executing a consensus pipeline on approximately 1.1 million facts resulted in a total API expenditure of approximately \$2,300. However, by releasing \textsc{OmniCSEval} as a comprehensive and reusable benchmark, we aim to offset this initial environmental and financial cost. Our framework is designed to prevent redundant and computationally expensive evaluations, ultimately saving compute for the broader NLP community.

\section*{Acknowledgement}
This research was supported in part by the National Natural Science Foundation of China (Grant Nos. 62276017, 62406033, U1636211, 61672081) and the State Key Laboratory of Complex \& Critical Software Environment (Grant No. SKLCCSE-2024ZX-18).

\bibliography{custom}

\appendix

\section{Source Dataset Details}
\label{sec:dataset_details}
\paragraph{SAMSum \citep{gliwa-etal-2019-samsum}.}
It comprises 16,369 dialogue-summary pairs designed to simulate real-life messenger conversations. Constructed by linguists, it captures informal linguistic features (e.g., slang, typos) across diverse daily topics.

\paragraph{DialogSum \citep{chen-etal-2021-dialogsum}.}
It consists of 13,460 dialogue-summary pairs covering diverse daily-life scenarios, including schooling, work, and shopping. Collected from spoken dialogue corpora (e.g., DailyDialog, DREAM), it captures face-to-face interactions typically between two speakers.

\paragraph{MediaSum \citep{zhu-etal-2021-mediasum}.}
This large-scale dataset comprises 463,596 dialogue-summary pairs derived from media interviews. Sourced from NPR and CNN transcripts, it covers open-domain topics such as politics, economy, and crime.

\paragraph{SummScreen \citep{chen-etal-2022-summscreen}.}
This dataset comprises 26,851 dialogue-summary pairs derived from TV series transcripts. It consists of two subsets: \textit{ForeverDreaming} and \textit{TVMegaSite}, covering diverse narrative genres such as drama and comedy. The dialogues feature complex, entity-centric communications with an average of over 20 speakers.

\paragraph{MovieSum \citep{saxena-keller-2024-moviesum}.}
It comprises 2,200 dialogue-summary pairs derived from movie screenplays. Collected from various script repositories, the screenplays are manually formatted using Celtx to preserve structural elements like scene headings and speaker utterances. It spans diverse genres such as drama, comedy, and action.

\paragraph{TweetSumm \citep{feigenblat-etal-2021-tweetsumm-dialog}.}
This dataset comprises 1,100 dialogues focusing on real-world customer service interactions. Sourced from the \textit{Customer Support on Twitter} corpus, it covers diverse domains such as airlines and retail.

\paragraph{TODSum \citep{zhao2021todsumtaskorienteddialoguesummarization}.}
This dataset comprises 9,906 dialogue-summary pairs derived from the MultiWOZ corpus, focusing on task-oriented interactions in customer service. It covers five domains: hotel, restaurant, attraction, taxi, and train.

\paragraph{ACI-Bench \citep{yim2023acibenchnovelambientclinical}.} This dataset comprises 207 dialogue-summary pairs focusing on doctor-patient interactions. Constructed through role-playing sessions involving medical experts, it includes both human and machine-generated transcripts of clinical encounters.

\paragraph{MTS-Dialog \citep{ben-abacha-etal-2023-empirical}.}
It comprises 1,700 dialogue-summary pairs focusing on doctor-patient interactions. The dialogues are manually simulated by healthcare professionals based on authentic clinical notes. It covers diverse specialties including neurology and orthopedics.

\paragraph{PriMock57 \citep{papadopoulos-korfiatis-etal-2022-primock57}.}
It comprises 57 dialogue-summary pairs focusing on mock primary care consultations. The dialogues feature simulated interactions between clinicians and patients, recorded via telemedicine software.

\paragraph{MeetingBank \citep{hu-etal-2023-meetingbank}.}
This benchmark comprises 6,892 dialogue-summary pairs derived from city council meetings across six U.S. cities. The dialogues are transcripts generated via ASR. It features multi-party interactions with 2 to 20 speakers covering public affairs.

\paragraph{QMSum \citep{zhong-etal-2021-qmsum}.}
It comprises 232 dialogue-summary pairs derived from professional meetings across three domains: \textit{Product}, \textit{Academic}, and \textit{Committee}. Sourced from the AMI and ICSI corpora as well as Welsh and Canadian parliamentary proceedings, it features extensive and complex multi-party interactions.

\begin{table*}[t]
    \centering
    \small
    \setlength{\abovecaptionskip}{8pt}
    \setlength{\tabcolsep}{5.0pt}
    \resizebox{\linewidth}{!}{
    \begin{tabular}{cccc cc cc}
        \toprule[1pt]
        \multirow{2.5}{*}{\textbf{Domain}} & \multirow{2.5}{*}{\textbf{Scenario}} & \multirow{2.5}{*}{\textbf{Dataset}} & \multirow{2.5}{*}{\textbf{Samples}} & \multicolumn{2}{c}{\textbf{Conversation}} & \multicolumn{2}{c}{\textbf{Reference}} \\
        \cmidrule(lr){5-6} \cmidrule(lr){7-8}
         & & & & \textbf{\#Tokens} & \textbf{\#Sentences} & \textbf{\#Tokens} & \textbf{\#Sentences} \\
        \midrule[0.5pt]
        Open-Ended & Daily Life & SAMSum & 150 & 212.4 & 20.6 & 36.0 & 2.6 \\
        Open-Ended & Daily Life & DialogSum & 150 & 217.4 & 19.0 & 40.9 & 2.3 \\
        Open-Ended & Media Interview & MediaSum-NPR & 150 & 864.3 & 42.4 & 49.8 & 2.4 \\
        Open-Ended & Media Interview & MediaSum-CNN & 150 & 2,500.6 & 139.5 & 44.6 & 2.1 \\
        Open-Ended & Screenplay & SummScreen-FD & 100 & 8,438.3 & 834.2 & 133.7 & 5.4 \\
        Open-Ended & Screenplay & SummScreen-TMS & 100 & 6,454.8 & 650.1 & 362.0 & 21.3 \\
        Open-Ended & Screenplay & MovieSum & 100 & 23,710.4 & 2,237.4 & 749.4 & 29.8 \\
        \midrule[0.5pt]
        Private & Customer Service & TweetSumm & 150 & 278.8 & 19.8 & 48.0 & 2.0 \\
        Private & Customer Service & TODSum & 150 & 254.1 & 24.3 & 64.3 & 4.1 \\
        Private & Healthcare & ACI-Bench & 130 & 1,416.4 & 90.4 & 545.5 & 34.5 \\
        Private & Healthcare & MTS-Dialog & 130 & 407.8 & 35.2 & 158.4 & 8.8 \\
        Private & Healthcare & PriMock57 & 40 & 2,123.0 & 179.6 & 211.6 & 11.4 \\
        Private & Meeting & MeetingBank & 75 & 9,905.7 & 638.0 & 106.3 & 2.9 \\
        Private & Meeting & QMSum-Product & 70 & 8,897.0 & 763.4 & 123.8 & 5.9 \\
        Private & Meeting & QMSum-Academic & 50 & 18,743.4 & 1,172.2 & 112.8 & 5.4 \\
        Private & Meeting & QMSum-Committee & 30 & 17,162.5 & 801.2 & 156.8 & 6.4 \\
        Private & Meeting & ECTSum & 75 & 3,475.4 & 125.9 & 93.7 & 5.3 \\
        \bottomrule[1pt]
    \end{tabular}
    }
    \caption{Detailed per-dataset statistics of our \textsc{OmniCSEval} benchmark.}
    \label{tab:sampled_data_details}
\end{table*}

\paragraph{ECTSum \citep{mukherjee-etal-2022-ectsum}.}
This dataset comprises 2,425 dialogue-summary pairs derived from earnings call transcripts collected from The Motley Fool. It captures financial discussions between company executives and analysts.

\section{Sampled Data Details}
\label{sec:sampled_data_details}
This section provides a granular breakdown of the 13 source datasets that constitute the \textsc{OmniCSEval} benchmark. Table \ref{tab:sampled_data_details} reports the specific sampling quotas and linguistic statistics for each dataset within their respective scenarios and domains.

\begin{figure}[t]
  \centering
  \setlength{\abovecaptionskip}{8pt}
  \includegraphics[width=0.999\linewidth]{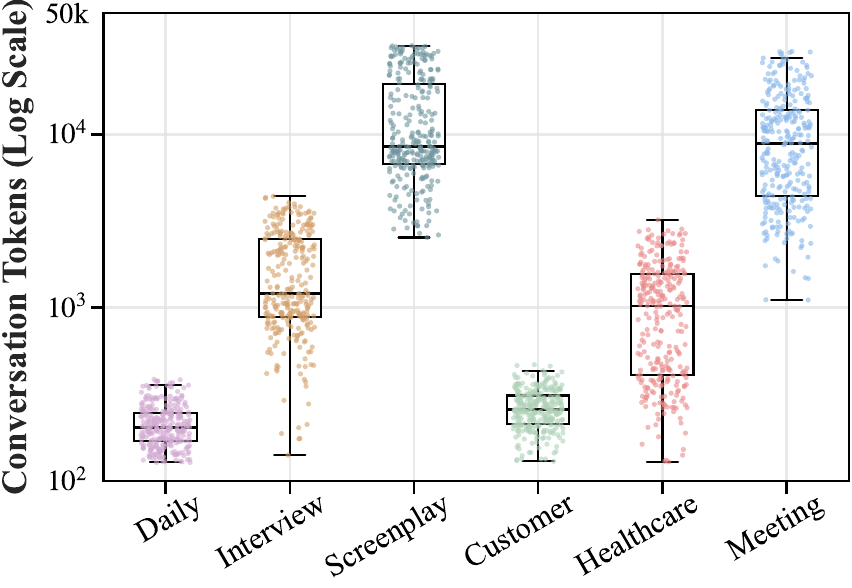}
  \caption{Conversation length distribution across scenarios. Each scatter point denotes an individual sample.}
  \label{fig:conversation_distribution}
\end{figure}

The sampling strategy primarily aims to maintain a balanced representation of 300 instances per scenario. However, for certain datasets with limited intrinsic sizes---specifically \textbf{PriMock57} (57 total), \textbf{QMSum-Academic} (59 total), and \textbf{QMSum-Committee} (36 total)---we made necessary sampling adjustments by sampling 40, 50, and 30 instances, respectively. These adjustments are essential to ensure that the sampled subsets remain representative of the source datasets while fulfilling the aggregate requirements for each scenario. In Figure~\ref{fig:conversation_distribution}, we present the length distribution of the resulting conversations across all scenarios.

\section{Evaluated LLM Details}
\label{sec:evaluated_llm_details}
This section provides detailed specifications for the 28 evaluated LLMs. As outlined in Table~\ref{tab:evaluated_llm_details}, our selection encompasses a diverse array of models, comprising 13 proprietary and 15 open-weight models across various parameter scales. For each model, we document key metadata, including the provider, version, parameter size, and maximum context window. We also specify the architectural type and the specific reasoning configuration (Effort / Budget). Additionally, Table~\ref{tab:pipeline_llm_details} details the 8 LLMs employed in our evaluation pipeline. To mitigate self-evaluation bias, these models are strictly isolated from the evaluated systems.

\begin{table*}[t]
    \centering
    \small
    \setlength{\abovecaptionskip}{8pt}
    \setlength{\tabcolsep}{6.0pt}
    \resizebox{\textwidth}{!}{
    \begin{tabular}{lccccccc}
        \toprule[1pt]
        \textbf{Model Name} & \textbf{Availability} & \textbf{Provider} & \textbf{Version} & \textbf{Size} & \textbf{Context} & \textbf{Type} & \textbf{Effort / Budget} \\
        \midrule[0.5pt]
        \rowcolor{gray!20}
        \multicolumn{8}{l}{\textit{Large Reasoning Language Models}} \\
       	\rule{0pt}{10pt}GPT-5 & Proprietary & OpenAI & 2025-08 & -- & 400k & Reasoning & Medium \\
       	Gemini 2.5 Pro & Proprietary & Google & 2025-06 & -- & 1,024k & Reasoning & 8,192 \\
        Qwen3-235B-A22B-Thinking & Open-Weights & Qwen & 2025-07 & 235B & 256k & Reasoning & 8,192 \\
        DeepSeek-R1 & Open-Weights & DeepSeek & 2025-05 & 671B & 160k & Reasoning & 8,192 \\
        GLM-4.6-Thinking & Open-Weights & Zhipu AI & 2025-09 & 335B & 200k & Hybrid & 8,192 \\
        Kimi-K2-Thinking & Open-Weights & MoonShot & 2025-11 & 1T & 256k & Reasoning & 8,192 \\
        MiniMax-M2 & Open-Weights & MiniMax & 2025-10 & 230B & 200k & Reasoning & 8,192 \\
        \midrule[0.5pt]
        \rowcolor{gray!20}
        \multicolumn{8}{l}{\textit{Small Reasoning Language Models}} \\
       	\rule{0pt}{10pt}GPT-5-mini & Proprietary & OpenAI & 2025-08 & -- & 400k & Reasoning & Medium \\
       	GPT-5-nano & Proprietary & OpenAI & 2025-08 & -- & 400k & Reasoning & Medium \\
       	Gemini 2.5 Flash & Proprietary & Google & 2025-06 & -- & 1,024k & Hybrid & 8,192 \\
       	Gemini 2.5 Flash Lite & Proprietary & Google & 2025-06 & -- & 1,024k & Hybrid & 8,192 \\
        Qwen3-14B-Thinking & Open-Weights & Qwen & 2025-04 & 14B & 128k & Hybrid & 8,192 \\
        Qwen3-8B-Thinking & Open-Weights & Qwen & 2025-04 & 8B & 128k & Hybrid & 8,192 \\
        GLM-Z1-9B & Open-Weights & Zhipu AI & 2025-04 & 9B & 128k & Reasoning & 8,192 \\
        \midrule[0.5pt]
        \rowcolor{gray!20}
        \multicolumn{8}{l}{\textit{Large Instruction Language Models}} \\
        \rule{0pt}{10pt}GPT-4.1 & Proprietary & OpenAI & 2025-04 & -- & 1,024k & Instruction & -- \\
        GPT-4o & Proprietary & OpenAI & 2024-11 & -- & 128k & Instruction & -- \\
        Qwen3-235B-A22B-Instruct & Open-Weights & Qwen & 2025-07 & 235B & 256k & Instruction & -- \\
        Qwen2.5-72B-Instruct-128K & Open-Weights & Qwen & 2024-09 & 72B & 128k & Instruction & -- \\
        DeepSeek-V3 & Open-Weights & DeepSeek & 2025-03 & 671B & 128k & Instruction & -- \\
        GLM-4.6-Instruct & Open-Weights & Zhipu AI & 2025-09 & 335B & 200k & Hybrid & Disable \\
        \midrule[0.5pt]
        \rowcolor{gray!20}
        \multicolumn{8}{l}{\textit{Small Instruction Language Models}} \\
        \rule{0pt}{10pt}GPT-4.1-mini & Proprietary & OpenAI & 2025-04 & -- & 1,024k & Instruction & -- \\
        GPT-4.1-nano & Proprietary & OpenAI & 2025-04 & -- & 1,024k & Instruction & -- \\
        GPT-4o-mini & Proprietary & OpenAI & 2024-07 & -- & 128k & Instruction & -- \\
       	Gemini 2.0 Flash & Proprietary & Google & 2025-02 & -- & 1,024k & Instruction & -- \\
       	Gemini 2.0 Flash Lite & Proprietary & Google & 2025-02 & -- & 1,024k & Instruction & -- \\
        Qwen3-14B-Instruct & Open-Weights & Qwen & 2025-04 & 14B & 128k & Hybrid & Disable \\
        Qwen3-8B-Instruct & Open-Weights & Qwen & 2025-04 & 8B & 128k & Hybrid & Disable \\
        GLM-4-9B-Chat & Open-Weights & Zhipu AI & 2024-06 & 9B & 128k & Instruction & -- \\
        \bottomrule[1pt]
    \end{tabular}
    }
    \caption{Detailed specifications of the evaluated LLMs. The \textbf{Type} column classifies the model's native architecture as \textit{Reasoning} (specialized for long chain-of-thought), \textit{Instruction} (standard generation without explicit reasoning), or \textit{Hybrid} (capable of both). \textbf{Effort / Budget} denotes the reasoning configuration applied in our benchmark, where \texttt{Disable} indicates that explicit reasoning was suppressed for Hybrid models.}
    \label{tab:evaluated_llm_details}
    \vspace{-5pt}
\end{table*}

\begin{table*}[t]
    \centering
    \small
    \setlength{\abovecaptionskip}{8pt}
    \setlength{\tabcolsep}{6.0pt}
    \resizebox{\textwidth}{!}{
    \begin{tabular}{lccccccc}
        \toprule[1pt]
        \textbf{Model Name} & \textbf{Availability} & \textbf{Provider} & \textbf{Version} & \textbf{Size} & \textbf{Context} & \textbf{Type} & \textbf{Effort / Budget} \\
        \midrule[0.5pt]
        Gemini 3 Pro & Proprietary & Google & 2025-11 & -- & 1,024k & Reasoning & High \\
        DeepSeek-V3.2-Instruct & Open-Weights & DeepSeek & 2025-12 & 671B & 160k & Hybrid & Disable \\
        GLM-4.7-Instruct & Open-Weights & Zhipu AI & 2025-12 & 355B & 200k & Hybrid & Disable \\
        Kimi-K2-Instruct & Open-Weights & MoonShot & 2025-09 & 1T & 256k & Instruction & -- \\
        Qwen3-Coder-480B-A35B & Open-Weights & Qwen & 2025-07 & 480B & 256k & Instruction & -- \\
        Qwen3-Next-80B-A3B-Instruct & Open-Weights & Qwen & 2025-09 & 80B & 256k & Instruction & -- \\
        Llama 4 Maverick & Open-Weights & Meta & 2025-04 & 400B & 1,024k & Instruction & -- \\
        Mistral Large 3 & Open-Weights & Mistral AI & 2025-12 & 675B & 256k & Instruction & -- \\
        \bottomrule[1pt]
    \end{tabular}
    }
    \caption{Detailed specifications of the 8 LLMs utilized within the \textsc{OmniCSEval} evaluation pipeline. To eliminate self-evaluation bias, these models are strictly isolated from the 28 systems being evaluated.}
    \label{tab:pipeline_llm_details}
\end{table*}

\section{Evaluation Metric Details}
\label{sec:formula_details}
\subsection{Forward Key Fact Matching}
\label{sec:fkfm}
Let $\mathcal{F}_{key} = \{f_1, f_2, \dots, f_N\}$ represent the set of key facts extracted from the source conversation, and $\mathcal{S} = \{s_1, s_2, \dots, s_M\}$ denote the set of sentences in the model-generated summary. We define an LLM-based matching function $\mathcal{M}(\cdot)$ that determines semantic entailment. The completeness score is calculated as the proportion of key facts that are successfully matched by the summary:
\[
    \mathrm{Completeness} = \frac{1}{N} \sum_{i=1}^{N} \mathbb{I}\left[ \exists s, \mathcal{M}(f_i, s) = 1 \right]
\]
where $\mathbb{I}[\cdot]$ is the indicator function, and $\mathcal{M}(f_i, s)=1$ if the judge model determines that summary sentence $s$ supports key fact $f_i$.

Similarly, the conciseness score is calculated as the density of informative summary sentences:
\[
    \mathrm{Conciseness} = \frac{1}{M} \sum_{j=1}^{M} \mathbb{I}\left[ \exists f, \mathcal{M}(f, s_j) = 1 \right]
\]

This metric measures the precision of summary sentences, penalizing sentences that do not contain salient information or cover redundant content.

\subsection{Backward Summary Fact Verification}
\label{sec:bsfv}

Let $\mathcal{A} = \{a_1, a_2, \dots, a_K\}$ denote the set of atomic facts decomposed from the model-generated summary, and $\mathcal{C}$ represent the source conversation context. We define an LLM-based verification function $\mathcal{V}(\cdot)$ to check factual consistency. The faithfulness score is calculated as the proportion of atomic facts supported by the source conversation:
\[
    \mathrm{Faithfulness} = \frac{1}{K} \sum_{k=1}^{K} \mathbb{I}\left[ \mathcal{V}(a_k, \mathcal{C}) = \text{supported} \right]
\]
where $\mathcal{V}(a_k, \mathcal{C})$ returns "supported" if the judge model finds evidence for atomic fact $a_k$ in the conversation context $\mathcal{C}$, and "not supported" otherwise.

\subsection{Meta-Evaluation Metrics}
\label{sec:meta_eval_metrics}
To measure the alignment between LLM-based fact-checkers and human annotations in our meta-evaluation (\S\ref{sec:meta_evaluation}), we employ Macro F1 for the key fact matching task and Balanced Accuracy (BAcc) for the summary fact verification task.

For the key fact matching task, the number of summary sentences varies across instances, resulting in dynamic dimensions for the binary matching matrices. To address this, we flatten the two-dimensional boolean matching maps of all evaluated key fact-summary sentence pairs into a unified one-dimensional list. Let $y \in \{0, 1\}$ represent the human-annotated ground truth and $\hat{y} \in \{0, 1\}$ denote the LLM prediction. The Macro F1 score is calculated as the unweighted mean of the $F_1$ scores for both the matched (1) and unmatched (0) classes:
\[
    \mathrm{Macro\ F1} = \frac{F_1^{(0)} + F_1^{(1)}}{2}
\]
where $F_1^{(c)}$ is the harmonic mean of precision and recall for class $c \in \{0, 1\}$.

For the summary fact verification task, which is formulated as a binary classification problem (supported vs. not supported), we utilize Balanced Accuracy to account for potential class imbalances in the test set. BAcc is defined as the arithmetic mean of the True Positive Rate (TPR) and the True Negative Rate (TNR):
\[
    \mathrm{BAcc} = \frac{\mathrm{TPR} + \mathrm{TNR}}{2}
\]
where $\mathrm{TPR}$ represents the proportion of actual supported facts correctly identified, and $\mathrm{TNR}$ represents the proportion of actual unsupported (hallucinated) facts correctly identified.

\section{Key Fact Details}
\label{sec:key_fact_details}
Table~\ref{tab:key-facts_statistics} reveals a clear tiered distribution of information volume across scenarios. Concise interactions, such as Daily Life and Customer Service, average fewer than 10 key facts, reflecting their focused nature. Conversely, the information load nearly doubles in structurally complex scenarios such as Screenplay and Meeting. Notably, Healthcare yields the highest fact density, containing the most key facts despite its moderate context length. This density stems from the high concentration of critical, discrete units inherent in clinical consultations, such as symptoms and treatment plans. Figure~\ref{fig:key-facts_examples} provides concrete examples.

\begin{table}[t]
    \centering
    \small
    \setlength{\abovecaptionskip}{8pt}
    \setlength{\tabcolsep}{4.5pt}
    \begin{tabular}{lcc}
        \toprule[1pt]
        \textbf{Scenario} & \textbf{\#Count (min-max)} & \textbf{\#Tokens (min-max)} \\
        \midrule[0.5pt]
        \rowcolor{gray!20}
        \multicolumn{3}{l}{\textit{Open-Ended}} \\
        Daily & 8.6 (4-16) & 102.8 (38-224) \\
        Interview & 15.1 (4-29) & 264.8 (47-484) \\
        Screenplay & 17.6 (10-31) & 318.9 (134-613) \\
        \midrule[0.5pt]
        \rowcolor{gray!20}
        \multicolumn{3}{l}{\textit{Private}} \\
        Customer & 9.9 (3-23) & 135.2 (38-272) \\
       	Healthcare & 21.4 (4-48) & 253.3 (40-496) \\
       	Meeting & 19.5 (10-47) & 392.7 (143-1,098) \\
        \midrule[0.5pt]
        \textbf{Overall} & 15.4 (3-48) & 244.6 (38-1,098) \\
        \bottomrule[1pt]
    \end{tabular}
    \caption{Statistics of key facts across different scenarios. The column \#Count denotes the average number of key facts per conversation, while \#Tokens represents the average total token count of these facts.}
    \label{tab:key-facts_statistics}
    \vspace{-5pt}
\end{table}

\section{Human Annotation Details}
\label{sec:human_annotation_details}
To reduce cognitive load and enhance annotation quality, we introduced assistive mechanisms. For the key fact matching task, we provided annotators with the summary sentences alongside their specific similarity scores to the key fact, highlighting the top three most similar sentences (or only the top one if the summary contained three or fewer sentences). For the summary fact verification task, we similarly employed context mapping by highlighting the top 25\scalebox{0.90}{$\%$} of source sentences most similar to the summary fact (capped at 2,048 tokens).

Throughout the annotation process, we did not explicitly provide any LLM pre-annotation results. However, when encountering challenging cases, annotators were permitted to access a designated verification platform\footnote{\url{https://openrouter.ai/chat}}. When utilizing this feature, annotators were required to consult the judgments of at least three distinct LLMs and synthesize these insights to reach a final, independent decision. Both tasks exhibited substantial inter-annotator agreement, achieving Fleiss' Kappa coefficients of 0.86 for key fact matching (computed by flattening the binary mapping matrices of key facts and summary sentences) and 0.63 for summary fact verification.

\section{Human Reference Analysis}

\begin{table}[t]
    \centering
    \small
    \setlength{\abovecaptionskip}{8pt}
    \setlength{\tabcolsep}{6pt}
    \begin{tabular}{lccc}
        \toprule[1pt]
        \textbf{Human Reference} & \textbf{Comp.} & \textbf{Conci.} & \textbf{Faith.} \\
        \midrule[0.5pt]
        Screenplay & 49.1 & 48.5 & 81.4 \\
        Media Interview & 17.7 & 71.6 & 76.9 \\
        Daily Life & 38.7 & 83.8 & 84.9 \\
        Meeting & 15.1 & 47.1 & 80.8 \\
        Healthcare & 81.8 & 71.6 & 84.6 \\
        Customer Service & 32.3 & 69.7 & 80.9 \\
        \rowcolor{MilkRed!70} \textbf{Overall} & \textbf{39.1} & \textbf{65.4} & \textbf{81.6} \\
        \midrule[0.5pt]
        \rowcolor{MilkGreen!70} \textbf{GPT-4.1-nano} (\textit{weakest}) & \textbf{42.6} & \textbf{70.1} & \textbf{93.4} \\
        \rowcolor{MilkGreen!70} \textbf{GPT-5} (\textit{best}) & \textbf{81.7} & \textbf{80.5} & \textbf{96.4} \\
        \bottomrule[1pt]
    \end{tabular}
    \caption{Performance of human-created reference summaries. Human summaries substantially underperform SOTA LLMs and even trail the weakest small model.}
    \label{tab:human_performance}
    \vspace{-5pt}
\end{table}

We further re-evaluate the original human-written reference summaries using the same bidirectional fact-checking framework. As shown in Table~\ref{tab:human_performance}, these references are substantially outperformed by frontier reasoning models, and even lag behind the weakest small instruction model across all three dimensions. This finding suggests that human references in many existing conversation summarization datasets should no longer be treated as reliable upper bounds for evaluating LLM-generated summaries. Many references are sparse, overly compressed, or shaped by dataset-specific annotation conventions, rather than by complete, concise, and faithful coverage of the source conversation. Therefore, reference-based similarity metrics such as ROUGE~\citep{lin2004rouge} and BERTScore~\citep{zhang2019bertscore} become increasingly inadequate: \textit{they may penalize high-quality LLM summaries that cover more key facts, while rewarding summaries that merely resemble incomplete references}.

The scenario-level results underscore further differences. Human references perform particularly poorly in Meeting and Media Interview, where they miss a large portion of key information. In contrast, Healthcare references remain relatively strong in completeness, highlighting the value of expert annotation and domain-specific summarization protocols. Nevertheless, even in Healthcare, human references still do not consistently surpass modern LLM outputs, reinforcing the need for source-grounded, fact-level evaluation rather than unconditional reliance on human references.

\section{Instructions}
\label{sec:instructions}
\begin{figure}[t]
  \centering
  \setlength{\abovecaptionskip}{5pt}
  \includegraphics[width=0.999\linewidth]{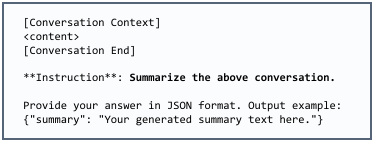}
  \caption{Conversation summarization instruction.}
  \label{fig:convo_summ_instruct}
  \vspace{0pt}
\end{figure}

\begin{figure}[t]
  \centering
  \setlength{\abovecaptionskip}{5pt}
  \includegraphics[width=0.999\linewidth]{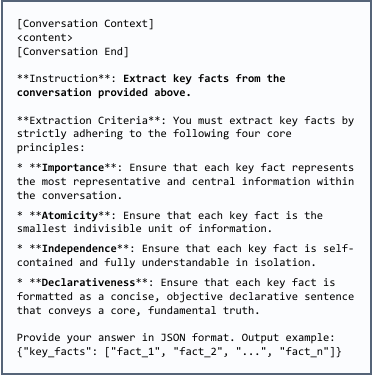}
  \caption{Key fact extraction instruction.}
  \label{fig:key-facts_extraction}
  \vspace{-5pt}
\end{figure}

In this section, we detail the complete set of instructions used to interact with the LLMs throughout our methodology. To ensure reproducibility, we provide the specific directives designed for each phase of our evaluation pipeline. These include the zero-shot conversation summarization task, alongside instructions for our human-LLM collaborative framework, such as key fact extraction, self-reflection, atomic fact decomposition, multi-LLM consensus filtering, and bidirectional fact-checking. A comprehensive overview and corresponding figure references are summarized in Table~\ref{tab:instruction_figures}.

\begin{table}[h]
    \centering
    \small
    \setlength{\abovecaptionskip}{8pt}
    \setlength{\tabcolsep}{6pt}
    \begin{tabular}{lc}
        \toprule[1pt]
        \textbf{Instruction} & \textbf{Reference} \\
        \midrule[0.5pt]
        Conversation Summarization & Figure~\ref{fig:convo_summ_instruct} \\
        Key Fact Extraction & Figure~\ref{fig:key-facts_extraction} \\
        Self-Reflection & Figure~\ref{fig:reflection_instruction} \\
        Summary Fact Decomposition & Figure~\ref{fig:atomic-facts_decomposition} \\
        Hallucinated Fact Filtering & Figure~\ref{fig:multi-llm-consensus} \\
        Key Fact Matching & Figure~\ref{fig:key-facts_verification} \\
        Summary Fact Verification & Figure~\ref{fig:ref-summ-facts_verification} \\
        \bottomrule[1pt]
    \end{tabular}
    \caption{Overview of the instructions.}
    \label{tab:instruction_figures}
    \vspace{0pt}
\end{table}

\begin{figure*}[t]
  \centering
  \setlength{\abovecaptionskip}{5pt}
  \includegraphics[width=0.999\linewidth]{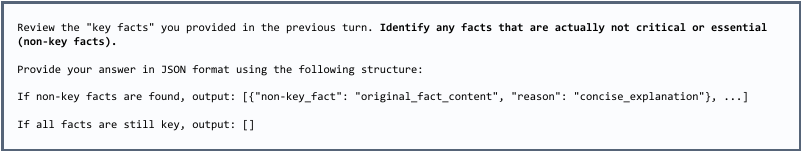}
  \caption{Self-reflection instruction.}
  \label{fig:reflection_instruction}
  \vspace{-5pt}
\end{figure*}

\begin{figure*}[t]
  \centering
  \setlength{\abovecaptionskip}{5pt}
  \includegraphics[width=0.999\linewidth]{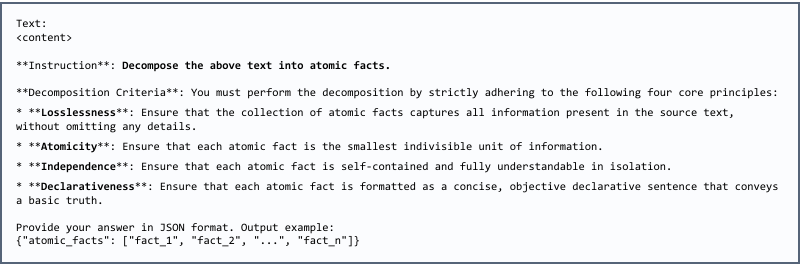}
  \caption{Summary fact decomposition instruction.}
  \label{fig:atomic-facts_decomposition}
  \vspace{-5pt}
\end{figure*}

\begin{figure*}[t]
  \centering
  \setlength{\abovecaptionskip}{5pt}
  \includegraphics[width=0.999\linewidth]{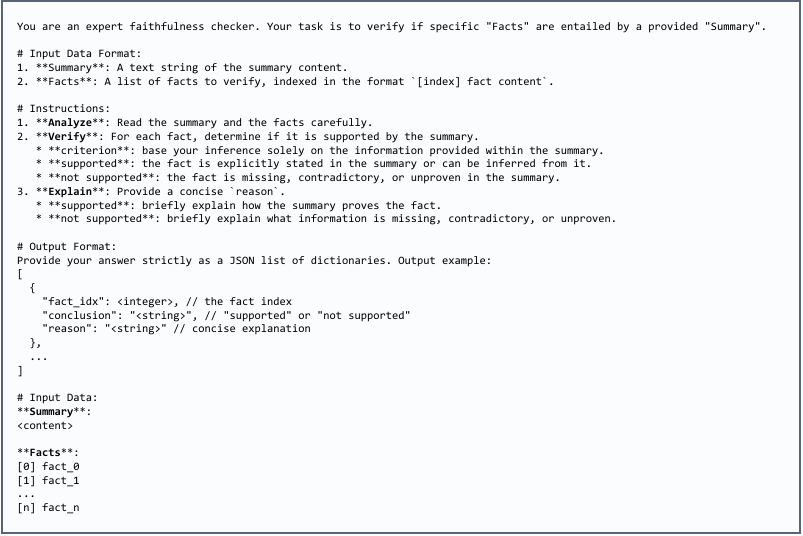}
  \caption{Hallucinated fact filtering instruction.}
  \label{fig:multi-llm-consensus}
  \vspace{-5pt}
\end{figure*}

\begin{figure*}[t]
  \centering
  \setlength{\abovecaptionskip}{5pt}
  \includegraphics[width=0.999\linewidth]{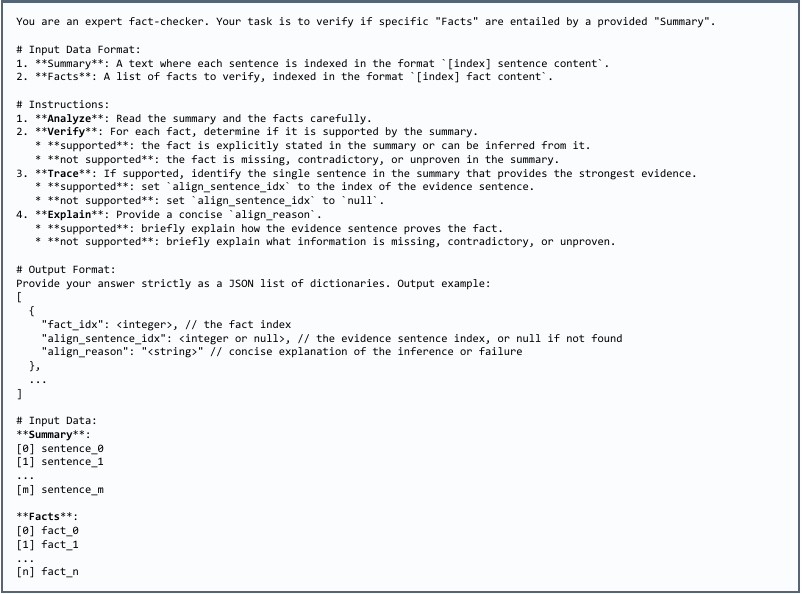}
  \caption{Key fact matching instruction.}
  \label{fig:key-facts_verification}
  \vspace{-5pt}
\end{figure*}

\begin{figure*}[h]
  \centering
  \setlength{\abovecaptionskip}{5pt}
  \includegraphics[width=0.999\linewidth]{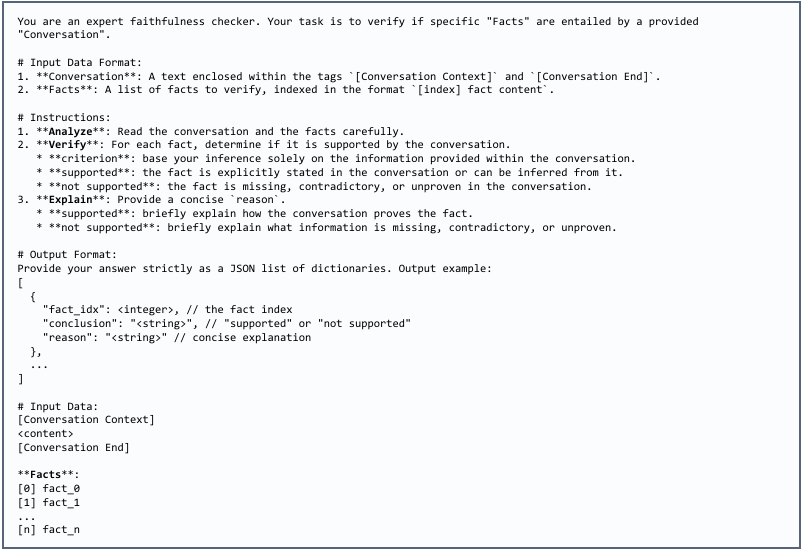}
  \caption{Summary fact verification instruction.}
  \label{fig:ref-summ-facts_verification}
  \vspace{-5pt}
\end{figure*}

\begin{figure*}[t]
  \centering
  \setlength{\abovecaptionskip}{8pt}
  \includegraphics[width=0.999\linewidth]{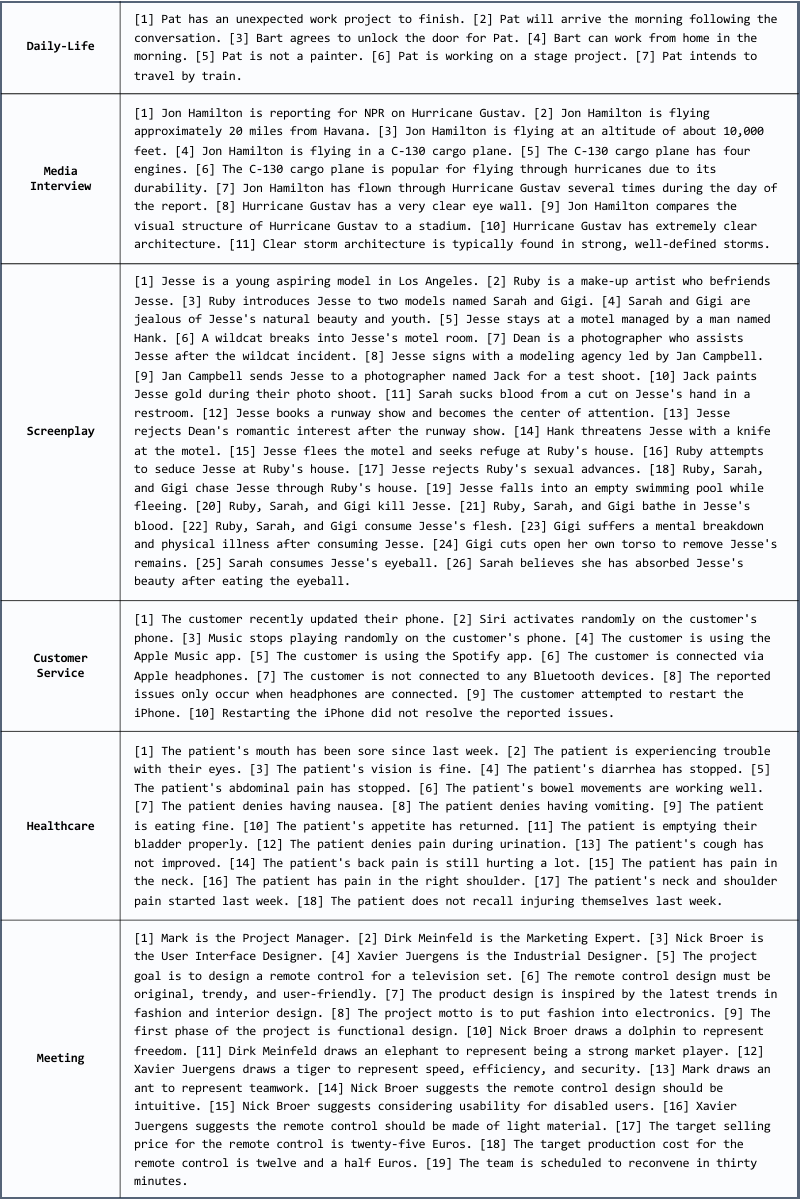}
  \caption{Examples of key facts across six scenarios.}
  \label{fig:key-facts_examples}
  \vspace{-5pt}
\end{figure*}

\begin{figure*}[t]
  \centering
  \setlength{\abovecaptionskip}{8pt}
  \includegraphics[width=0.999\linewidth]{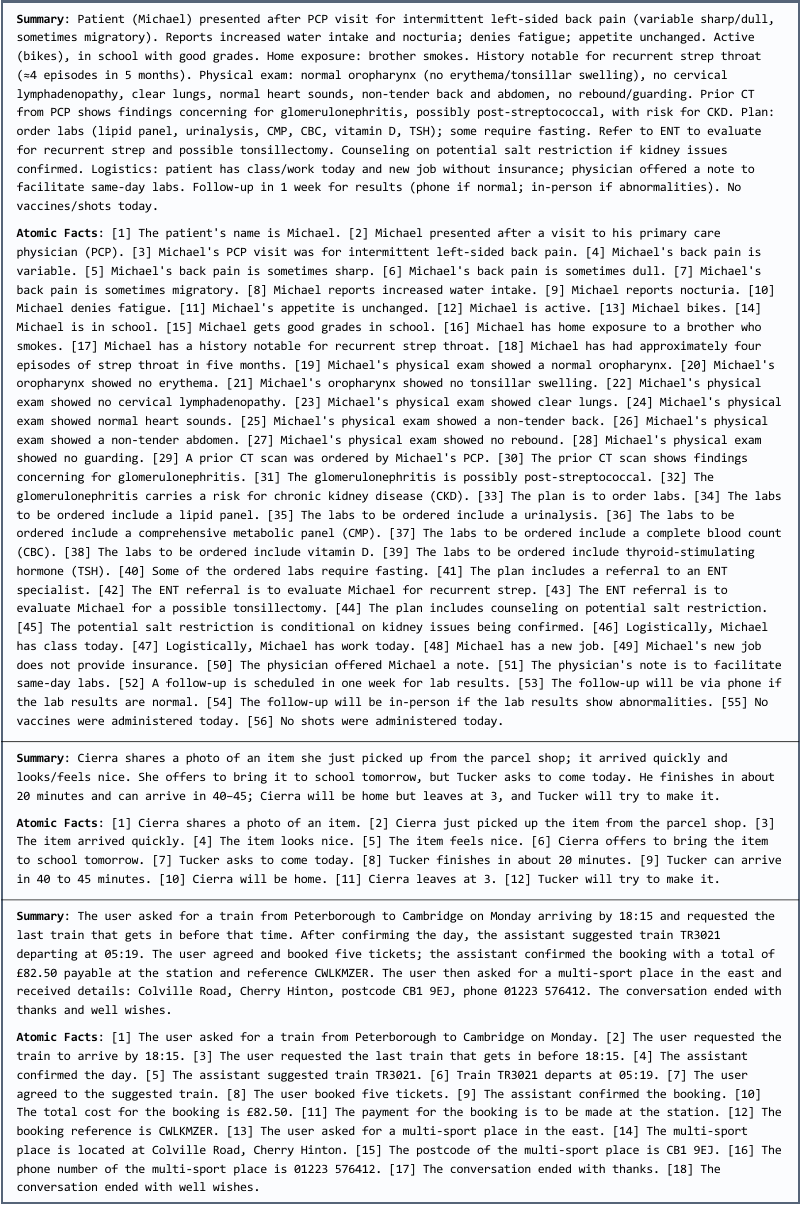}
  \caption{Examples of decomposed summary facts.}
  \label{fig:summary-facts_examples}
  \vspace{-5pt}
\end{figure*}

\end{document}